\documentclass[10pt,twocolumn,letterpaper]{article}

\usepackage[pagenumbers]{configs/cvpr}      

\usepackage{amsmath,amsfonts,bm}

\def\eqref#1{equation~\ref{#1}}

\def\1{\bm{1}}

\def\eps{{\epsilon}}

\def\vs{{\bm{s}}}

\DeclareMathAlphabet{\mathsfit}{\encodingdefault}{\sfdefault}{m}{sl}
\SetMathAlphabet{\mathsfit}{bold}{\encodingdefault}{\sfdefault}{bx}{n}

\usepackage[table]{xcolor}
\definecolor{citecolor}{HTML}{0071BC}
\definecolor{linkcolor}{HTML}{ED1C24}
\usepackage[pagebackref=false, breaklinks=true, letterpaper=true, colorlinks, bookmarks=false,  citecolor=citecolor, linkcolor=linkcolor, urlcolor=gray]{hyperref}
\usepackage{orcidlink}
\usepackage{natbib}
\usepackage{amsfonts}
\usepackage{url}
\usepackage{graphicx}
\usepackage{wrapfig}
\usepackage{tabularx}
\usepackage{color, colortbl}
\usepackage{physics}
\usepackage{booktabs}
\usepackage{multirow}
\usepackage{xcolor}
\usepackage{graphicx}
\usepackage{footnote}
\usepackage{tabularx}
\usepackage{float}
\usepackage{lipsum}
\usepackage{multicol}
\usepackage{siunitx}
\usepackage{tabularx,booktabs}
\usepackage{etoolbox}
\usepackage{algorithm}
\usepackage{listings}
\usepackage{booktabs}
\usepackage{threeparttable}
\usepackage{subcaption}
\usepackage{enumitem}
\usepackage{adjustbox}
\usepackage{textpos}  
\usepackage{bm} 
\usepackage{pgffor}  

\usepackage[british,english,american]{babel}

\usepackage{etoolbox}
\makeatletter
\AfterEndEnvironment{algorithm}{\let\@algcomment\relax}
\AtEndEnvironment{algorithm}{\kern2pt\hrule\relax\vskip3pt\@algcomment}
\let\@algcomment\relax
\newcommand\algcomment[1]{\def\@algcomment{\footnotesize#1}}
\renewcommand\fs@ruled{\def\@fs@cfont{\bfseries}\let\@fs@capt\floatc@ruled
  \def\@fs@pre{\hrule height.8pt depth0pt \kern2pt}%
  \def\@fs@post{}%
  \def\@fs@mid{\kern2pt\hrule\kern2pt}%
  \let\@fs@iftopcapt\iftrue}
\makeatother

\definecolor{deemph}{gray}{0.6}

\definecolor{LightCyan}{rgb}{0.88,1,1}
\definecolor{LightRed}{rgb}{1,0.5,0.5}
\definecolor{LightYellow}{rgb}{1,1,0.88}
\definecolor{Grey}{rgb}{0.75,0.75,0.75}
\definecolor{DarkGrey}{rgb}{0.55,0.55,0.55}
\definecolor{DarkGreen}{rgb}{0,0.65,0}

\def\paperID{xxxx} 
\def\confName{xxxx}
\def\confYear{xxxx}

\newlength\savewidth\newcommand\shline{\noalign{\global\savewidth\arrayrulewidth
  \global\arrayrulewidth 1pt}\hline\noalign{\global\arrayrulewidth\savewidth}}

\newcommand\midline{\noalign{\global\savewidth\arrayrulewidth
  \global\arrayrulewidth 0.5pt}\hline\noalign{\global\arrayrulewidth\savewidth}}

\newcommand{\tablestyle}[2]{\setlength{\tabcolsep}{#1}\renewcommand{\arraystretch}{#2}\centering\footnotesize}
\definecolor{baselinecolor}{gray}{.9}

\newcolumntype{x}[1]{>{\centering\arraybackslash}p{#1pt}}
\newcolumntype{y}[1]{>{\raggedright\arraybackslash}p{#1pt}}
\newcolumntype{z}[1]{>{\raggedleft\arraybackslash}p{#1pt}}

\usepackage{xspace}
\makeatletter
\DeclareRobustCommand\onedot{\futurelet\@let@token\@onedot}
\def\@onedot{\ifx\@let@token.\else.\null\fi\xspace}
\def\eg{\emph{e.g}\onedot} 
\def\ie{\emph{i.e}\onedot} 
 
 \def\vs{\emph{vs}\onedot}

\makeatother

\colorlet{darkgreen}{green!65!black}
\colorlet{darkblue}{blue!75!black}
\colorlet{darkred}{red!80!black}
\definecolor{lightblue}{HTML}{0071bc}
\definecolor{lightgreen}{HTML}{39b54a}

\newcommand{\x}{\bm{x}}
\newcommand{\e}{\bm{\epsilon}}
\renewcommand{\v}{\bm{v}}
\newcommand{\z}{\bm{z}}
\newcommand{\net}{\text{\texttt{net}}}

\renewcommand{\paragraph}[1]{\vspace{1.25mm}\noindent\textbf{#1}}

\newcommand{\app}{\raise.17ex\hbox{$\scriptstyle\sim$}}

{
\setlength{\leftmargini}{9pt}%
\begin{itemize}%
\setlength{\itemsep}{0pt}%
\setlength{\topsep}{0pt}%
\setlength{\partopsep}{0pt}%
\setlength{\parskip}{0pt}}%
{\end{itemize}}

\newcommand{\hhs}{\hspace{-0.001em}}
\newcommand{\vvs}{\vspace{-.1em}}

\graphicspath{{./figures/}}

\begin{document}

\title{
Back to Basics: Let Denoising Generative Models Denoise
}

\author{
Tianhong Li \qquad Kaiming He \vspace{.5em} \\
MIT
\vspace{1.em}
}

\maketitle

\begin{abstract}
Today's denoising diffusion models do not ``denoise'' in the classical sense, \ie, they do not directly predict clean images. 
Rather, the neural networks predict noise or a noised quantity.
In this paper, we suggest that predicting clean data and predicting noised quantities are fundamentally different.
According to the manifold assumption, natural data should lie on a low-dimensional manifold, whereas noised quantities do not. With this assumption, we advocate for models that directly predict clean data, which allows apparently under-capacity networks to operate effectively in very high-dimensional spaces. We show that simple, large-patch Transformers on pixels can be strong generative models: using no tokenizer, no pre-training, and no extra loss. Our approach is conceptually nothing more than ``\mbox{\textbf{Just image Transformers}}'', or \textbf{JiT}, as we call it. We report competitive results using JiT with large patch sizes of 16 and 32 on ImageNet at resolutions of 256 and 512, where predicting high-dimensional noised quantities can fail catastrophically. 
With our networks mapping back to the basics of the manifold,
our research goes back to basics and pursues a self-contained paradigm for Transformer-based diffusion on raw natural data.

\end{abstract}

\section{Introduction}
\label{sec:intro}

When diffusion generative models were first developed \cite{SohlDickstein2015,Song2019,Ho2020}, the core idea was supposed to be \mbox{\textit{denoising}}, \ie, \textit{predicting a clean image} from its corrupted version. 
However, two significant milestones in the evolution of diffusion models turned out to deviate from the goal of directly predicting clean images. 
First, predicting the noise itself (known as ``$\e$-prediction'') \cite{Ho2020} made a pivotal difference in generation quality and largely popularized these models. Later, diffusion models were connected to flow-based methods \cite{lipman2022flow,liu2022flow,albergo2023building} through predicting the flow velocity (``\mbox{$\v$-prediction}'' \cite{salimans2022progressive}), a quantity that combines clean data and noise.
Today, diffusion models in practice commonly predict noise or a noised quantity (\eg, velocity).

\begin{figure}[t]
\centering
\includegraphics[width=0.8\linewidth]{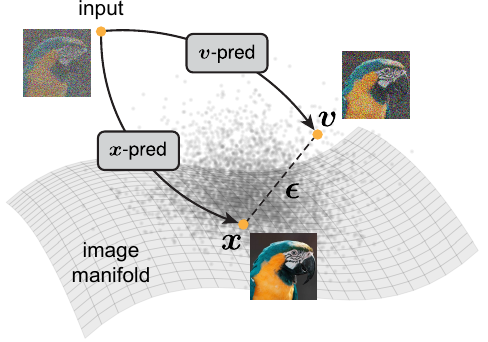}
\vspace{-.5em}
\caption{
\textbf{The Manifold Assumption} \cite{Chapelle2006SemiSupervised}
hypothesizes that natural images lie on a low-dimensional manifold within the high-dimensional pixel space.
While a clean image $\x$ can be modeled as on-manifold, the noise $\e$ or flow velocity $\v$ (\eg, $\v = \x - \e$) is inherently off-manifold. Training a neural network to 
predict a clean image (\ie, $\x$-prediction) is \textit{fundamentally different} from training it to predict noise or a noised quantity (\ie, $\e$/$\v$-prediction).
}
\label{fig:teaser}
\vspace{-1em}
\end{figure}

Extensive studies \cite{salimans2022progressive,Karras2022edm,hoogeboom2023simple,esser2024scaling} have shown that predicting the clean image (``\mbox{$\x$-prediction}" \cite{salimans2022progressive}) is closely related to $\e$- and $\v$-prediction, provided that the \textit{weighting} of the prediction loss is reformulated accordingly (detailed in \cref{sec:analysis}).
Owing to this relation, less attention has been paid to what the network should directly predict, implicitly assuming that the network is capable of performing the assigned task.

However, the roles of clean images and a noised quantity (including the noise itself) are \textit{not} equal.
In machine learning, it has long been hypothesized \cite{Chapelle2006SemiSupervised,carlsson2009topology} that
\textit{``(high-dimensional) data lie (roughly) on a low-dimensional manifold''} (\cite[p.6]{Chapelle2006SemiSupervised}).
Under this manifold assumption, while clean data can be modeled as lying on a low-dimensional manifold, a noised quantity is inherently distributed across the full high-dimensional space \cite{vincent2010stacked} (see \cref{fig:teaser}). \textit{Predicting clean data is fundamentally different from predicting noise or a noised quantity.}

Consider a scenario where a low-dimensional manifold is embedded in a high-dimensional observation space.
Predicting noise in this high-dimensional space requires \textit{high capacity}: the network needs to preserve all information about the noise.
In contrast, a \textit{limited-capacity} network can still predict the clean data, as it only needs to retain the low-dimensional information while filtering out the noise.

When a low-dimensional space (\eg, image latent \cite{Rombach2022}) is used, 
the difficulty of predicting noise is alleviated, yet at the same time is \textit{hidden} rather than solved.
When it comes to pixel or other high-dimensional spaces, existing diffusion models can still struggle to address the \textit{curse of dimensionality} \cite{Chapelle2006SemiSupervised}.
The heavy reliance on a pre-trained latent space  prevents diffusion models from being self-contained.

In pursuit of a self-contained principle, there has been strong focus on advancing diffusion modeling in the pixel space \cite{chen2023importance,hoogeboom2023simple,hoogeboom2024simpler,chen2025pixelflow,wang2025pixnerd}. 
In general, these methods explicitly or implicitly avoid the information bottleneck in the networks, \eg, by using dense convolutions or smaller patches, increasing channels, or adding long skip connections. 
We suggest that these designs may stem from the demand to predict high-dimensional noised quantities.

In this paper, we return to first principles and let the neural network \textit{directly} predict the clean image. By doing so, we show that a \textit{plain} Vision Transformer (ViT) \cite{Dosovitskiy2021}, operating on large image patches consisting of raw pixels, can  be effective for diffusion modeling.
Our approach is \textit{self-contained} and does not rely on any pre-training or auxiliary loss --- \textit{no latent tokenizer \cite{Rombach2022}, no adversarial loss \cite{goodfellow2014generative,Rombach2022}, no perceptual loss \cite{zhang2018unreasonable,Rombach2022} (thus no pre-trained classifier \cite{simonyan2014very}), and no representation alignment \cite{repa} (thus no self-supervised pre-training \cite{oquab2023dinov2})}. 
Conceptually, our model is nothing more than ``Just image Transformers'', or \textbf{JiT}, as we call it, applied to diffusion.

We conduct experiments on the ImageNet dataset \cite{deng2009imagenet} at resolutions of 256 and 512, using JiT models with patch sizes 16 and 32 respectively. 
Even though the patches are very high-dimensional (hundreds or thousands), our models using $\x$-prediction can easily produce strong results, where $\e$- and $\v$-prediction fail catastrophically.  Further analysis shows that it is \mbox{unnecessary} for the network width to match or exceed the patch dimension; in fact, and surprisingly, a \textit{bottleneck} design can even be beneficial, echoing observations from classical manifold learning.

Our effort marks a step toward a self-contained ``Diffusion\,+\,Transformer'' philosophy \cite{Peebles2023} on native data. Beyond computer vision, this philosophy is highly desirable in other domains involving natural data (\eg, proteins, molecules, or weather), where a tokenizer can be difficult to design. By minimizing domain-specific designs, we hope that the general ``Diffusion\,+\,Transformer'' paradigm originated from computer vision will find broader applicability.

\section{Related Work}

\paragraph{Diffusion Models and Their Predictions.} The pioneering work on diffusion models \cite{SohlDickstein2015} proposed to learn a reversed stochastic process, in which the network predicts the parameters of a normal distribution (\eg, mean and standard deviation). Five years after its introduction, this method was revitalized and popularized by Denoising Diffusion Probabilistic Models (DDPM) \cite{Ho2020}: a \textit{pivotal} discovery was to make noise the prediction target (\ie, $\e$-prediction). 

The relationships among different prediction targets were then investigated in \cite{salimans2022progressive} (originally in the context of model distillation), where the notion of $\v$-prediction was also introduced. Their work focused on the weighting effects introduced by reparameterization.

Meanwhile, EDM \cite{Karras2022edm} reformulated the diffusion problem around a \textit{denoiser function}, which constitutes a major milestone in the evolution of diffusion models. However, EDM adopted a \textit{pre-conditioned} formulation, where the direct output of the network is \textit{not} the denoised image. While this formulation is preferable in low-dimensional scenarios, it still inherently requires the network to output a quantity that mixes data and noise (more comparisons in appendix).

Flow Matching models \cite{lipman2022flow,liu2022flow,albergo2023building} can be interpreted as a form of $\v$-prediction \cite{salimans2022progressive} within the diffusion modeling framework. 
Unlike pure noise, $\v$ is a combination of data and noise.
The connections between flow-based models and previous diffusion models have been established \cite{esser2024scaling}.
Today, diffusion models and their flow-based counterparts are often studied under a unified framework.

\paragraph{Denoising Models.} Over the past decades, the concept of denoising has been closely related to representation learning.
Classical methods, exemplified by BM3D \cite{dabov2007image} and others \cite{portilla2003image,elad2006image,zoran2011learning}, leveraged the assumptions of sparsity and low dimensionality to perform image denoising.

Denoising Autoencoders (DAEs) \cite{vincent2008extracting,vincent2010stacked} were developed as an unsupervised representation learning method, using denoising as their training objective. They leveraged the manifold assumption \cite{Chapelle2006SemiSupervised} (\cref{fig:teaser}) to learn meaningful representations that approximate the low-dimensional data manifold. DAEs can be viewed as a form of Denoising Score Matching \cite{vincent2011connection}, which in turn is closely related to modern score-based diffusion models \cite{Song2019,Song2021}. Nevertheless, while it is natural for DAEs to predict clean data for manifold learning, in score matching, predicting the score function effectively amounts to predicting the noise (up to a scaling factor), \ie, $\e$-prediction.

\paragraph{Manifold Learning.} Manifold learning has been a classical field \cite{roweis2000nonlinear,tenenbaum2000global} focused on learning low-dimensional, nonlinear representations from observed data. In general, manifold learning methods leverage \textit{bottleneck} structures \cite{tishby2000information,rifai2011contractive,makhzani2013k,alemi2016deep} that encourage only useful information to pass through. Several studies have explored the connections between manifold learning and generative models \cite{loaiza2024deep,humayun2024secrets,chen2024deconstructing}. Latent Diffusion Models (LDMs) \cite{Rombach2022} can be viewed as manifold learning in the first stage via an autoencoder, followed by diffusion in the second stage.

\paragraph{Pixel-space Diffusion.} 
While latent diffusion \cite{Rombach2022} has become the default choice in the field today, the development of diffusion models originally began with pixel-space formulations \cite{Song2019,Ho2020,nichol2021improved,dhariwal2021diffusion}.
Early pixel-space diffusion models were typically based on dense convolutional networks, most often a U-Net \cite{ronneberger2015u}. These models often use over-complete channel representations (\eg, transforming $H{\times}W{\times}3$ into $H{\times}W{\times}128$ in the first layer), accompanied by long-range skip connections. While these models work well for $\epsilon$- and $v$-prediction, their dense convolutional structures are typically computationally expensive.
Applying these convolutional models to high-resolution images does not lead to catastrophic degradation, and as such, research in this direction has commonly focused on noise schedules and/or weighting schemes \cite{chen2023importance,hoogeboom2023simple,hoogeboom2024simpler,Kingma2023} to further improve generation quality.

In contrast, applying a Vision Transformer (ViT) \cite{Dosovitskiy2021} directly to pixels presents a more challenging task. \mbox{\textit{Standard}} ViT architectures adopt an aggressive patch size (\eg, 16$\times$16 pixels), resulting in a high-dimensional token space that can be comparable to, or larger than, the Transformer's hidden dimension. 
SiD2 \cite{hoogeboom2024simpler} and PixelFlow \cite{chen2025pixelflow} adopt hierarchical designs that start from smaller patches; however, these models are ``FLOP-heavy'' \cite{hoogeboom2024simpler} and lose the inherent generality and simplicity of standard Transformers. \mbox{PixNerd} \cite{wang2025pixnerd} adopts a NeRF head \cite{mildenhall2021nerf} that integrates information from the Transformer output, noisy input, and spatial coordinates, with training further assisted by representation alignment \cite{repa}.

Even with these special-purpose designs, the architectures in these works typically start from the ``L'' (Large) or ``XL'' size.
In fact, a latest work \cite{yao2025reconstruction} suggests that a large hidden size appears necessary for high dimensionality.

\paragraph{High-dimensional Diffusion.} 
When using ViT-style architectures, modern diffusion models are still challenged by high-dimensional input spaces, whether in pixels or latents.
In the literature \cite{chen2024deconstructing,yao2025reconstruction,shi2025latent}, it has been \textit{repeatedly} reported that ViT-style diffusion models degrade rapidly and \textit{catastrophically} when the per-token dimensionality increases, regardless of the use of pixels or latents.

Concurrently with our work, a line of research 
\cite{zheng2025diffusion,lei2025advancing,shi2025latent} resorts to \textit{self-supervised pre-training} to address high-dimensional diffusion.
In contrast to these efforts, we show that high-dimensional diffusion is achievable \textit{without} any pre-training, and using \textit{just} Transformers.

\paragraph{$\x$-prediction.} 
The formulation of $\x$-prediction is natural and not new; it can be traced back at least to the original DDPM \cite{Ho2020} (see their code \cite{ho2020diffusion_code}). However, DDPM observed that $\e$-prediction was substantially better, which later became the go-to solution.
In later works (\eg, \cite{hoogeboom2024simpler}), although the analysis was sometimes preferably conducted in the $\x$-space, the actual prediction was often made in other spaces, likely for legacy reasons.

When it comes to the image restoration application addressed by diffusion \cite{delbracio2023inversion,xie2023diffusion,milanfar2025denoising}, it is natural for the network to predict the clean data, as this is the ultimate goal of image restoration. 
Concurrent with our work, \cite{hafner2025training} also advocates the use of $\x$-prediction, for generative world models that are conditional on previous frames.

Our work does not aim to reinvent this fundamental concept; rather, we aim to draw attention to a largely overlooked yet critical issue in the context of high-dimensional data with underlying low-dimensional manifolds.

\definecolor{xcol}{RGB}{255,240,210}    
\definecolor{epscol}{RGB}{210,230,255}  
\definecolor{vcol}{RGB}{210,245,210}    

\newcommand{\mathhl}[2]{
  \tikz[baseline=(X.base)] \node[fill=#1, rounded corners=1pt, inner sep=1pt] (X)
  {$\displaystyle #2$};
}

\newcommand{\hlx}[1]{\mathhl{xcol}{#1}}
\newcommand{\hleps}[1]{\mathhl{epscol}{#1}}
\newcommand{\hlv}[1]{\mathhl{vcol}{#1}}

\begin{table*}[t]
\vspace{-1em}
\centering
{
\tablestyle{2.5pt}{1.2}
\small
\begin{tabular}{l l | x{90} | x{90} | x{90} } 
& & \textbf{(a) $\x$-pred } & \textbf{(b) $\e$-pred} & \textbf{(c) $\v$-pred} \\
 & & \makebox[5em]{ $\hlx{\x_\theta}:=\net_\theta(\z_t, t)$} & \makebox[5em]{$\hleps{\e_\theta}:=\net_\theta(\z_t, t)$} & \makebox[5em]{$\hlv{\v_\theta}:=\net_\theta(\z_t, t)$} \\
\shline
\textbf{~(1)~$\x$-loss:} & $\mathbb{E} \|\x_\theta - \x \|^2$ &
$\hlx{\x_\theta}$ &
$\x_\theta=(\bm{z}_t{-}(1{-}t)\hleps{\e_\theta}) / t$
&
$\x_\theta= (1{-}t)\hlv{\v_\theta}{+} \bm{z}_t$ \\
\midline
\textbf{~(2)~$\e$-loss:} &  $\mathbb{E}\|\hspace{1pt}\e_\theta\hspace{0.5pt} - \hspace{0.75pt}\e\hspace{1pt} \|^2$ &
$\e_\theta=(\bm{z}_t{-} t\hlx{\x_\theta}) / (1 - t)$
&
$\hleps{\e_\theta}$ &
$\e_\theta=\bm{z}_t{-} t\hlv{\v_\theta}$ \\
\midline
\textbf{~(3)~$\v$-loss:} & $\mathbb{E}\|\hspace{0.5pt}\v_\theta\hspace{0.25pt} - \hspace{0.25pt}\v\hspace{0.5pt} \|^2$ &
$\v_\theta= (\hlx{\x_\theta}{-}\bm{z}_t) / (1 - t)$
&
$\v_\theta= ( \bm{z}_t{-} \hleps{\e_\theta} ) / t$ 
&
$\hlv{\v_\theta}$ \\
\end{tabular}
}
\vspace{-.75em}
\caption{
\textbf{All possible combinations} of defining the loss and network prediction in $\x$, $\v$, or $\e$ spaces. The direct network outputs are highlighted in colors. For any off-diagonal entry where the network output space differs from the loss space, a transformation on the network output is applied.
}
\label{tab:xev}
\vspace{-.75em}
\end{table*}

\section{On Prediction Outputs of Diffusion Models}
\label{sec:analysis}

Diffusion models can be formulated in the space of $\x$, $\e$, or $\v$. The choice of the space determines not only where the loss is defined, but also what the network predicts.
Importantly, \textit{the loss space and the network output space need not be the same}. This choice can make critical differences.

\subsection{Background: Diffusion and Flows}

Diffusion models can be formulated from the perspective of ODEs \cite{chen2018neural,Song2021,lipman2022flow, song2020denoising}. We begin our formulation with the flow-based paradigm \cite{lipman2022flow,liu2022flow,albergo2023building}, \ie, in the $\v$-space, as a simpler starting point, and then discuss other spaces.

Consider a data distribution $\x \sim p_\text{data}(\x)$ and a noise distribution $\e \sim p_\text{noise}(\e)$ (\eg, $\e\sim\mathcal{N}(0,\mathbf{I})$). During training, a noisy sample $\z_t$ is an interpolation: $\z_t\,=\,a_t\,\x\,+\,b_t\,\e$,
where $a_t$ and $b_t$ are pre-defined noise schedules at time $t \in [0, 1]$. In this paper, we use a linear schedule \cite{lipman2022flow,liu2022flow,albergo2023building}\footnotemark: $a_t = t$ and $b_t = 1 - t$. This gives:
\begin{equation}
\z_t = t \, \x + (1 - t)\, \e \label{eq:z},
\end{equation}
which leads to $\z_t\sim p_\text{data}$ when $t{=}1$.
We use the logit-normal distribution over $t$ \cite{esser2024scaling}, \ie, $\text{logit}(t) \sim \mathcal{N}(\mu,\sigma^2)$.

\footnotetext{Our analysis in this paper is applicable to other schedules.}

A flow velocity $\v$ is defined as the time-derivative of $\z$, that is, $\v_t = \z_t' = a'_t \x + b'_t \e$. Given \cref{eq:z}, we have:
\begin{equation}
\v = \x - \e.
\label{eq:v}
\end{equation}
The flow-based methods \cite{lipman2022flow,liu2022flow,albergo2023building} minimize a loss function defined as:
\begin{equation}
\mathcal{L} = \mathbb{E}_{t,\x,\e} \|  \v_\theta(\z_t, t) - \v \|^2,
\label{eq:loss_v}
\end{equation}
where $\v_\theta$ is a function parameterized by $\theta$. While $\v_\theta$ is often the \textit{direct} output of a network $\v_\theta\,=\,\net_\theta(\z_t, t)$ \cite{lipman2022flow,liu2022flow,albergo2023building}, it can also be a transform of it, as we will elaborate.

Given the function $\v_\theta$, sampling is done by solving an ordinary differential equation (ODE) for $\z$ \cite{lipman2022flow,liu2022flow,albergo2023building}:
\begin{equation}
{{d\z_t} / {d t}} = \v_\theta(\z_t, t),
\label{eq:ode}
\end{equation}
starting from $\z_0 \sim p_\text{noise}$ and ending at $t=1$. In practice, this ODE can be approximately solved using numerical solvers. By default, we use a 50-step Heun.

\subsection{Prediction Space and Loss Space}

\paragraph{Prediction Space.}
The network's direct output can be defined in any space: $\v$, $\x$, or $\e$. Next, we discuss the resulting transformation. Note that in the context of this paper, we refer to it as ``$\x$, $\e$, $\v$-\textbf{prediction}'', \textit{only} when the network $\net_\theta$'s \textit{direct} output is strictly $\x$, $\e$, $\v$, respectively.

Given three unknowns ($\x$, $\e$, $\v$) and one network output, we require two additional constraints to determine all three unknowns. The two constraints are given by \cref{eq:z} and (\ref{eq:v}).
\textit{For example}, when we let the direct network output $\net_\theta$ be $\x$, we solve the following set of equations:
\vspace{-3pt}
\begin{equation}
\left\{
\begin{aligned}
\x_\theta &= \net_\theta \\
\z_t &= t \, \x_\theta + (1 - t)\, \e_\theta \\
\v_\theta &= \x_\theta - \e_\theta
\end{aligned}
\right.
\label{eq:set_x}
\vspace{-3pt}
\end{equation}
Here, the notations $\x_\theta$, $\e_\theta$, and $\v_\theta$ suggest that they are all predictions dependent on $\theta$. Solving this equation set gives:
$\e_\theta = (\z_t - t  \x_\theta) / (1 - t)$ and $\v_\theta = (\x_\theta - \z_t) / (1 - t)$, that is, both $\e_\theta$ and $\v_\theta$ can be computed from $\z_t$ and the network $\x_\theta$.
These are summarized in \cref{tab:xev} in column \textbf{(a)}.

Similarly, when we let the direct network output $\net_\theta$
be $\e$ or $\v$, we obtain the other sets of equations
(by replacing the first one in \cref{eq:set_x}). The transformations are summarized in \cref{tab:xev} in the columns of \textbf{(b)}, \textbf{(c)} for $\e$-, $\v$-prediction.
This shows that when one quantity in $\{\x, \e, \v \}$ is predicted, the other two can be inferred. The derivations in many prior works (\eg, \cite{salimans2022progressive,esser2024scaling}) are special cases covered in \cref{tab:xev}.

\paragraph{Loss Space.}
While the loss is often defined in one reference space (\eg, $\v$-loss in \cref{eq:loss_v}), 
conceptually, one can define it in any space. 
It has been shown \cite{salimans2022progressive,esser2024scaling} that with a given reparameterization from one prediction space to another, the loss is effectively reweighted.

\textit{For example}, consider the combination of $\x$\textbf{-prediction} and $\v$\textbf{-loss} in \cref{tab:xev}\textbf{(3)(a)}. We have
$\v_\theta = (\x_\theta - \z_t)/(1 - t)$
as the prediction and
$\v = (\x - \z_t) / (1-t)$
as the target. The $\v$-loss in \cref{eq:loss_v} becomes: 
$
\mathcal{L} = \mathbb{E} \|  \v_\theta(\z_t, t) - \v \|^2 = \mathbb{E} \frac{1}{(1-t)^2} \| \x_\theta(\z_t, t) - \x \|^2,
$
which is a \textit{reweighted} form of the $\x$-loss. A transformation like this one can be done for any prediction space and any loss space listed in \cref{tab:xev}. 

Put together, consider the three \textit{unweighted} losses defined in $\{\x, \e, \v \}$, and the three forms of the network direct output, there are \textit{nine possible combinations} (\cref{tab:xev}).
Each combination constitutes a valid formulation, and \textit{no} two among the nine cases are mathematically equivalent.

\paragraph{Generator Space.}
Regardless of the combination used, to perform generation at inference-time,
we can always transform the network output to the $\v$-space (\cref{tab:xev}, row (3)) and solve the ODE in \cref{eq:ode} for sampling. As such, all nine combinations are legitimate generators.

\begin{figure}[t]
\vspace{-1em}
\centering
\tablestyle{1.5pt}{1.7}
\begin{tabular}{@{}>{\centering\arraybackslash}m{0.10\linewidth}
                >{\centering\arraybackslash}m{0.21\linewidth}
                >{\centering\arraybackslash}m{0.21\linewidth}
                >{\centering\arraybackslash}m{0.21\linewidth}
                >{\centering\arraybackslash}m{0.21\linewidth}@{}}
& {ground-truth} &
  \textbf{$\x$-pred} &
  \textbf{$\e$-pred} &
  \textbf{$\v$-pred} \\[3pt]

\textbf{\scriptsize $D{=}2$} &
\includegraphics[width=\linewidth]{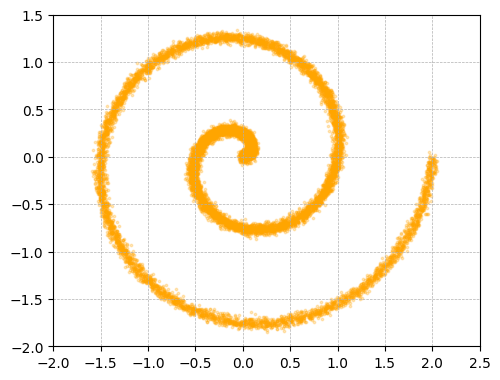} &
\includegraphics[width=\linewidth]{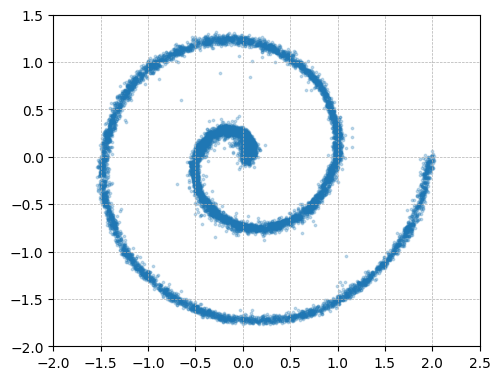} &
\includegraphics[width=\linewidth]{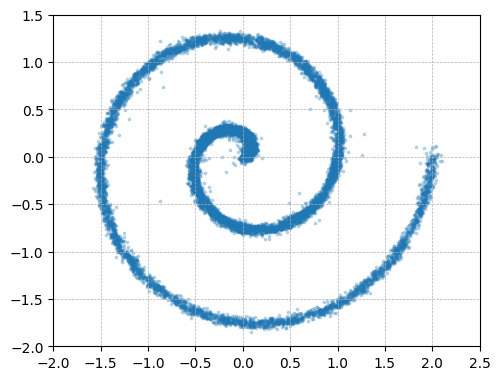} &
\includegraphics[width=\linewidth]{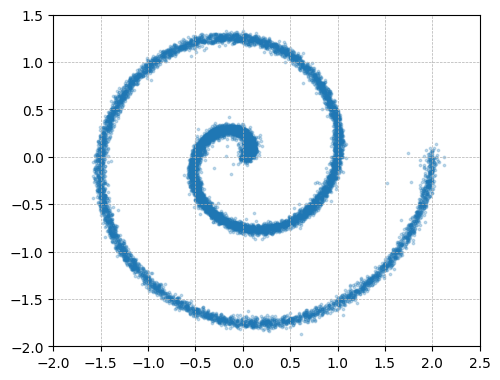} \\[8pt]

\textbf{\scriptsize $D{=}8$} &
\includegraphics[width=\linewidth]{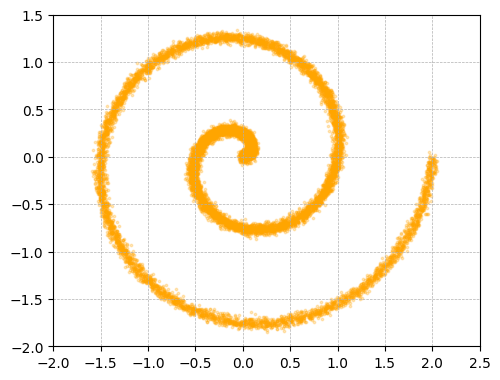} &
\includegraphics[width=\linewidth]{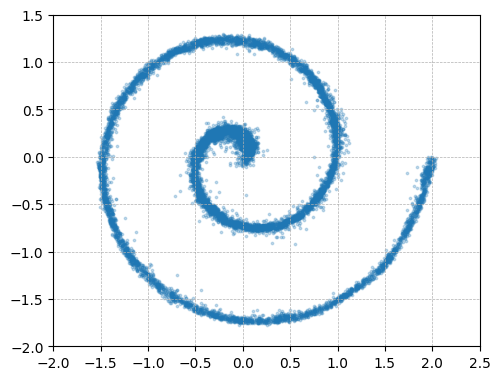} &
\includegraphics[width=\linewidth]{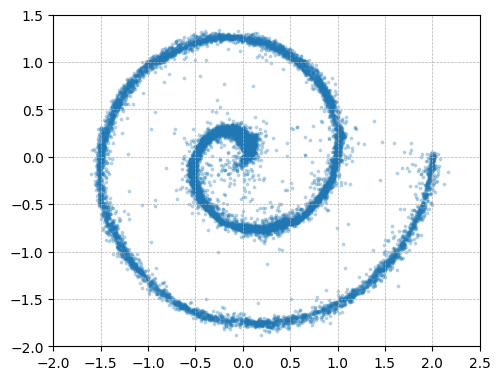} &
\includegraphics[width=\linewidth]{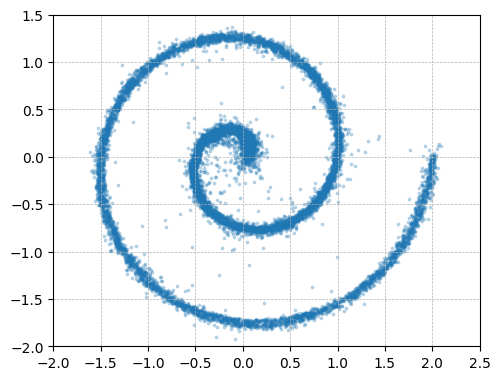} \\[8pt]

\textbf{\scriptsize $D{=}16$ } &
\includegraphics[width=\linewidth]{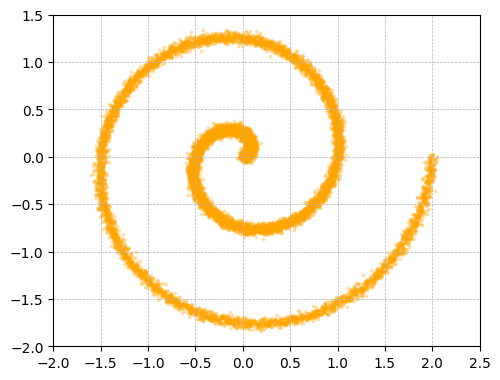} &
\includegraphics[width=\linewidth]{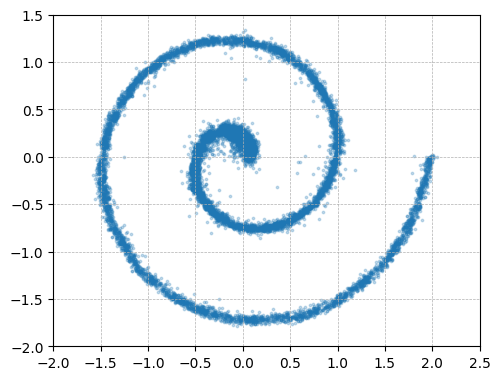} &
\includegraphics[width=\linewidth]{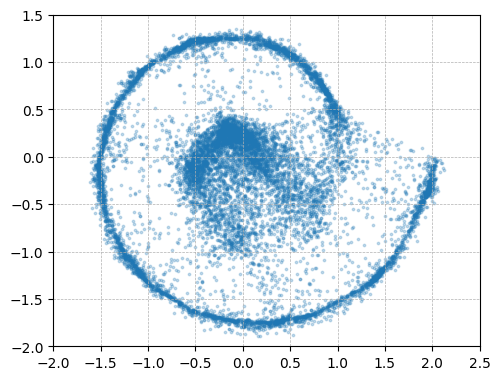} &
\includegraphics[width=\linewidth]{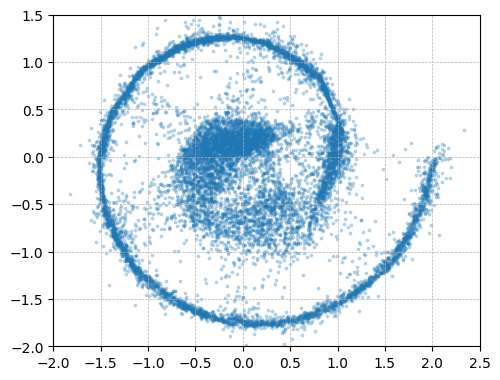} \\[8pt]

\textbf{\scriptsize $D{=}512$} &
\includegraphics[width=\linewidth]{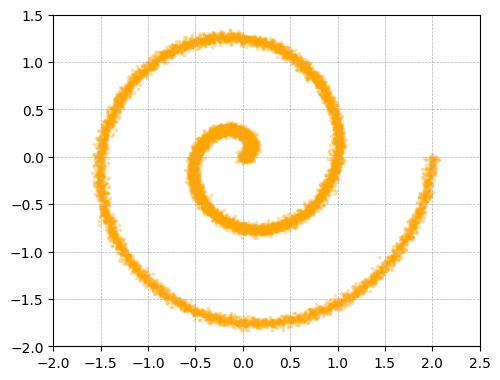} &
\includegraphics[width=\linewidth]{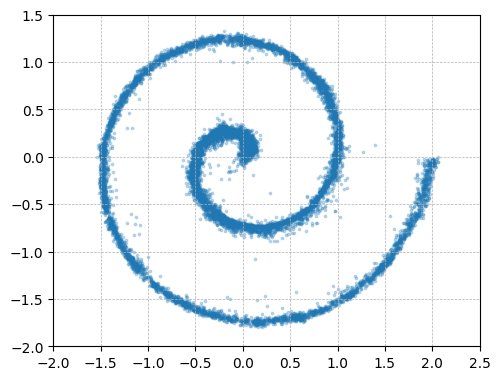} &
\includegraphics[width=\linewidth]{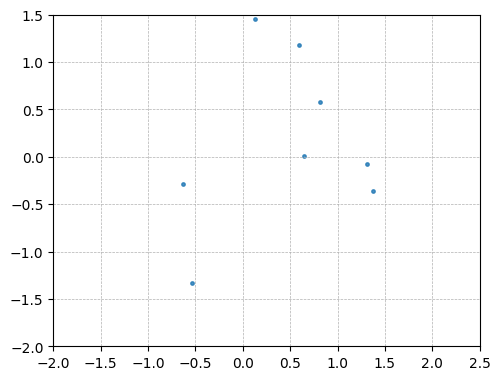} &
\includegraphics[width=\linewidth]{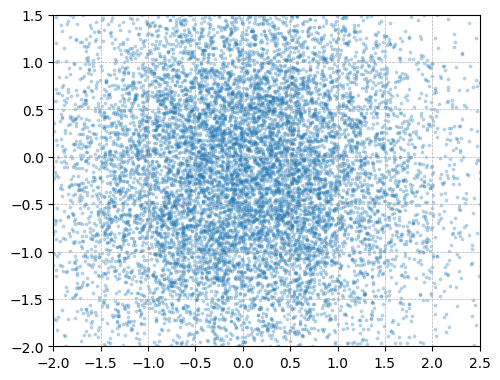} \\
\end{tabular}
\vspace{-1em}
\caption{
\textbf{Toy Experiment}: $d$-dimensional ($d=2$) underlying data is ``{buried}'' in a $D$-dimensional space, by a fixed, random, column-orthogonal projection matrix.
In the $D$-dim space, we train a simple generative model ({5-layer ReLU MLP with 256-dim hidden units}). The projection matrix is {unknown} to the model, and we only use it for visualizing the output.
In this toy experiment, \textit{with the observed dimension $D$ increasing, only \textbf{$\x$-prediction} can produce reasonable results.
}
}
\vspace{-1em}
\label{fig:toy}
\end{figure}

\subsection{Toy Experiment}

According to the manifold assumption \cite{Chapelle2006SemiSupervised}, the data $\x$ tends to lie in a low-dimensional manifold 
(\cref{fig:teaser}), while noise $\e$ and velocity $\v$ are off-manifold. Letting a network directly predict the clean data $\x$ should be more tractable. We verify this assumption in a toy experiment in this section.

We consider the toy case of $d$-dimensional underlying data ``buried'' in an observed $D$-dimensional space ($d < D$). We synthesize this scenario using a projection matrix $P \in \mathbb{R}^{D \times d}$ that is column-orthogonal, \ie, $P^{\top} P = I_{d \times d}$. This matrix $P$ is randomly created and fixed. The observed data is $\x = P \hat{\x} \in \mathbb{R}^{D}$, where the underlying data is $\hat{\x} \in \mathbb{R}^{d}$. The matrix $P$ is unknown to the model, and as such, it is a \mbox{$D$-dimensional} generation problem for the model. 

We train a {5-layer ReLU MLP with 256-dim hidden units} as the generator and visualize the results in \cref{fig:toy}. 
We obtain these visualizations by projecting the $D$-dim generated samples back to $d$-dim using $P$. We investigate the cases of $D \in \{2, 8, 16, 512\}$ for $d=2$.
We study $\x$, $\e$, or $\v$-prediction, all using the $\v$-loss, \ie, \cref{tab:xev}\textbf{(3)(a-c)}.

\cref{fig:toy} shows that {\textit{only {\textbf{$\x$-prediction}} can produce reasonable results when $D$ increases}}. For $\e$-/$\v$-prediction, the models struggle at $D{=}16$, and fail catastrophically when $D{=}512$, where the \mbox{256-dim} MLP is under-complete.

Notably, {\textit{{\textbf{$\x$-prediction}} can work well even when the model is {under-complete}}}. Here, the 256-dim MLP inevitably discards information in the $D{=}512$-dim space. However, since the true data is in a low-dimensional $d$-dim space, $\x$-prediction can still perform well, as the ideal output is implicitly $d$-dim. We draw similar observations in the case of real data on ImageNet, as we show next.

\definecolor{bad}{RGB}{255,235,235}    
\definecolor{good}{RGB}{235,255,235}    

\newcommand{\exphl}[2]{%
  \tikz[baseline=(X.base)] \node[fill=#1, rounded corners=1pt, inner sep=3pt] (X)
  {$\displaystyle #2$};%
}

\newcommand{\hlbad}[1]{\exphl{bad}{#1}}
\newcommand{\hlgood}[1]{\exphl{good}{#1}}

\section{``Just Image Transformers'' for Diffusion}

Driven by the above analysis, we show that \textit{plain} Vision Transformers (ViT) \cite{Dosovitskiy2021} operating on pixels can work surprisingly well, simply using $\x$-prediction.

\subsection{Just Image Transformers}

The core idea of ViT \cite{Dosovitskiy2021} is ``\textit{Transformer on Patches} (ToP)''\footnotemark.
Our architecture design follows this philosophy.

\footnotetext{Quoting Lucas Beyer.}

\begin{figure}[t]
\centering
\includegraphics[width=0.6\linewidth]{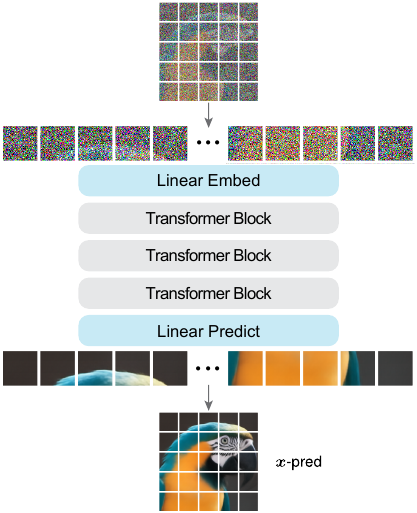}
\vspace{-.3em}
\caption{\textbf{The ``Just image Transformer'' (JiT) architecture}:
simply a plain ViT \cite{Dosovitskiy2021} on patches of pixels for $\x$-prediction.
}
\vspace{-.8em}
\label{fig:arch}
\end{figure}

Formally, consider $H{\times}W{\times}C$-dim image data ($C{=}3$). All $\x$, $\e$, $\v$ and $\z_t$ share this same dimensionality. 
Given an image, we divide it into non-overlapping patches of size $p{\times}p$, resulting in a sequence of a length $\frac{H}{p}{\times}\frac{W}{p}$. Each patch is a $p{\times}p{\times}3$-dim vector. This sequence is processed by a linear embedding projection, added with positional embedding \cite{vaswani2017attention}, and mapped by a stack of Transformer blocks \cite{vaswani2017attention}. 
The output layer is a linear predictor that projects each token back to a $p{\times}p{\times}3$-dim patch.
See \cref{fig:arch}.

As standard practice, the architecture is conditioned on time $t$ and a given class label. We use adaLN-Zero \cite{Peebles2023} for conditioning and will discuss other options later.
Conceptually, this architecture amounts to the Diffusion Transformer (DiT) \cite{Peebles2023} directly applied to patches of pixels.

The overall architecture is nothing more than ``Just image Transformers'', which we refer to as \textbf{JiT}. For example, we investigate \textbf{JiT/16} (\ie, patch size $p{=}16$, \cite{Dosovitskiy2021}) on $256{\times}256$ images, and \textbf{JiT/32} ($p{=}32$)  on $512{\times}512$ images. These settings respectively result in a dimensionality of 768 ($16{\times}16{\times}3$) and 3072 ($32{\times}32{\times}3$) per patch. 
Such high-dimensional patches can be handled by $\x$-prediction.

\subsection{What to Predict by the Network?}

We have summarized the nine possible combinations of loss space and prediction space in \cref{tab:xev}.
For each of these combinations, we train a ``Base'' \cite{Dosovitskiy2021} model (JiT-B), which has a hidden size of 768-dim per token.
We study JiT-B/16 at 256$\times$256 resolution in \cref{tab:3x3s}(a). As a reference, we examine \mbox{JiT-B/4} (\ie, 
$p{=}4$) at 64$\times$64 in \cref{tab:3x3s}(b). 
In both settings, the sequence length is the same (16${\times}$16).

We draw the following observations:

\DeclareRobustCommand{\bggood}[1]{%
  \begingroup
  \setlength{\fboxsep}{1pt}%
  \raisebox{0pt}[\height][0pt]{\colorbox{good}{#1}}%
  \endgroup
}
\DeclareRobustCommand{\bgbad}[1]{%
  \begingroup
  \setlength{\fboxsep}{1pt}%
  \raisebox{0pt}[\height][0pt]{\colorbox{bad}{#1}}%
  \endgroup
}

\paragraph{$\x$-prediction is critical.}
In \cref{tab:3x3s}(a) with JiT-B/16, {\textit{only $\x$-prediction performs well}}, and it works across all three losses. 
Here, a patch is 768-d ($16{\times}16{\times}3$), which coincides with the hidden size of 768 in \mbox{JiT-B}. While this may seem ``about enough'', in practice the models may require additional capacity, \eg, to handle positional embeddings.
For $\e$-/$\v$-prediction, the model does not have enough capacity to separate and retain the noised quantities. 
These observations are similar to those in the toy case (\cref{fig:toy}).

As a comparison, we examine JiT-B/4 at 64$\times$64 resolution (\cref{tab:3x3s}(b)). Here, all cases perform reasonably well: the accuracy gaps among the nine combinations are marginal, not decisive. The dimensionality is 48 ($4{\times}4{\times}3$) per patch, well below the hidden size of 768 in \mbox{JiT-B}, which explains why all combinations work reasonably well. We note that many previous \textit{latent} diffusion models have a similarly small input dimensionality and therefore were not exposed to the issue we discuss here.

\begin{table}[t]
\vspace{-1em}
\centering
\tablestyle{2.5pt}{1.2}
\begin{tabular}{x{45} | x{45} |  x{45} |  x{45} } 
 & \textbf{$\x$-pred}  & \textbf{$\e$-pred} & \textbf{$\v$-pred} \\
\shline
\textbf{$\x$-loss} & \cellcolor{good} 10.14 & \cellcolor{bad} 379.21 & \cellcolor{bad} 107.55 \\
\midline
\textbf{$\e$-loss} & \cellcolor{good} 10.45 & \cellcolor{bad} 394.58 & \cellcolor{bad} 126.88 \\
\midline
\textbf{$\v$-loss} & \cellcolor{good} \hspace{2pt} {8.62} & \cellcolor{bad} 372.38 & \cellcolor{bad} \hspace{2pt} 96.53 \\
\multicolumn{4}{c}{\textbf{(a)} ImageNet 256$\times$256, \textbf{JiT-B/16}} \\
\end{tabular}
\\
\vspace{.5em}
\tablestyle{2.5pt}{1.2}
\begin{tabular}{x{45} | x{45} | x{45} | x{45} } 
 & \textbf{$\x$-pred}  & \textbf{$\e$-pred} & \textbf{$\v$-pred} \\
\shline
\textbf{$\x$-loss} & \cellcolor{good} 5.76 & \cellcolor{good} 6.20 & \cellcolor{good} 6.12 \\
\midline
\textbf{$\e$-loss} & \cellcolor{good} 3.56 & \cellcolor{good} 4.02 & \cellcolor{good} 3.76 \\
\midline
\textbf{$\v$-loss} & \cellcolor{good} 3.55 & \cellcolor{good} 3.63 & \cellcolor{good} 3.46 \\
 \multicolumn{4}{c}{\textbf{(b)} ImageNet 64$\times$64, \textbf{JiT-B/4}}
\end{tabular}
\vspace{-.5em}
\caption{\textbf{Results of all combinations} of loss space and network space (see \cref{tab:xev}), evaluated by FID-50K on ImageNet:
\textbf{(a)} JiT-B/16 at 256 resolution, 768-d per patch;
\textbf{(b)} JiT-B/4 at 64 resolution, 48-d per patch.
We annotate {\bgbad{catastrophic failures}} in red and \bggood{reasonable results} by green.
Settings: 200 epochs, with CFG \cite{ho2022classifier}.
}
\label{tab:3x3s}
\vspace{-.2em}
\end{table}

\paragraph{Loss weighting is not sufficient.} Our work is not the first to enumerate the combinations of relevant factors. In \cite{salimans2022progressive}, it has explored the combinations of loss \textit{weighting} and network predictions. Their experiments were done on the low-dimensional CIFAR-10 dataset, using a U-net. Their observations were closer to ours on ImageNet 64$\times$64.\footnotemark{}

\footnotetext{In the CIFAR-10 experiments in \cite{salimans2022progressive}, they enumerated three types of loss weighting and three types of network outputs. In 8 out of these 9 combinations, their models work reasonably well (see their Tab.~1).
}

However, \cref{tab:3x3s}(a) on ImageNet 256$\times$256 suggests that \textit{loss weighting is not the whole story}.
On one hand, both $\e$- and $\v$-prediction fail catastrophically in \cref{tab:3x3s}(a), regardless of the loss space, which corresponds to different effective weightings in different loss spaces (as discussed).
On the other hand, $\x$-prediction works across \textit{all} three loss spaces: the loss weighting induced by the $\v$-loss is preferable, but not critical.\footnotemark{}

\footnotetext{From \cref{tab:xev}(a), we see that with $\x$-prediction, the coefficients of $\x_\theta$ are $1$, ${t}{/}{(1{-}t)}$, and $-{1}{/}(1{-}t)$ in the three rows. 
When converting to $\x$-loss, the weights of $\x$-loss are $1$, ${t^2}{/}{(1{-}t)^2}$, and ${1}{/}{(1{-}t)^2}$, respectively.}

\paragraph{Noise-level shift is not sufficient.} Prior works \cite{chen2023importance,hoogeboom2023simple,hoogeboom2024simpler} have suggested that increasing the noise level is useful for high-resolution pixel-based diffusion. We examine this in \cref{tab:mu} with JiT-B/16. 
As we use the logit-normal distribution \cite{esser2024scaling} for sampling $t$ (see appendix), the noise level can be shifted by changing the parameter $\mu$ of this distribution: intuitively, shifting $\mu$ towards the negative side results in smaller $t$ and thus increases the noise level (\cref{eq:z}).

\cref{tab:mu} shows that when the model already performs decently (here, \textbf{$\x$-pred}), appropriately high noise is beneficial, which is consistent with prior observations \cite{chen2023importance,hoogeboom2023simple,hoogeboom2024simpler}.
However, adjusting the noise level \textit{alone} cannot remedy $\e$- or $\v$-prediction: their failure stems inherently from the inability to propagate high-dimensional information.

As a side note, according to \cref{tab:mu}, we set $\mu$ {=} --0.8 in other experiments on ImageNet 256$\times$256.

\paragraph{Increasing hidden units is not necessary.} Since the capacity can be limited by the network width (\ie, numbers of hidden units), a natural idea is to increase it. 
However, this remedy is neither principled nor feasible when the observed dimension is very high. We show that this is not necessary in the case of $\x$-prediction.

In \cref{tab:in1024} and \cref{tab:scale} in the next section, we show results of \textbf{JiT/32} at resolution 512 and \textbf{JiT/64} at resolution 1024, 
using a \textit{proportionally large} patch size of $p{=}32$ or $p{=}64$.
This amounts to 3072-dim (\ie, $32{\times}32{\times}3$) or 12288-dim
per patch, substantially larger than the hidden size of B, L, and H models (defined in \cite{Dosovitskiy2021}).
Nevertheless, $\x$-prediction works well; in fact, it works \textit{without} any modification other than scaling the noise proportionally (\eg, by 2$\times$ and 4$\times$ at resolution 512 and 1024; see appendix).

This evidence suggests that the network design can be largely \textit{decoupled} from the observed dimensionality, as is the case in many other neural network applications. Increasing the number of hidden units can be beneficial (as widely observed in deep learning), but it is not decisive.

\begin{table}[t]
\vspace{-1em}
\centering
\tablestyle{4pt}{1.2}
\begin{tabular}{z{40} x{16} | x{45} | x{45} | x{45}}
\multicolumn{2}{r|}{$t$-shift ($\mu$)}   & $\x$\textbf{-pred} & $\e$\textbf{-pred} & $\v$\textbf{-pred} \\
\shline
{\scriptsize(lower noise)} & \textcolor{black!0}{--}0.0   & \cellcolor{good} \hspace{2pt} 14.44 & \cellcolor{bad} 464.25 & \cellcolor{bad} 120.03 \\
& --0.4 & \cellcolor{good} \hspace{6pt} 9.79  & \cellcolor{bad} 372.91 & \cellcolor{bad} 109.93 \\
& --0.8 & \cellcolor{good} \hspace{6pt} \textbf{8.62} & \cellcolor{bad} 372.36 & \cellcolor{bad} \hspace{2pt} 96.53 \\
{\scriptsize(higher noise)} & --1.2 & \cellcolor{good} \hspace{6pt} 8.99  & \cellcolor{bad} 355.25 & \cellcolor{bad} 106.85 \\
\end{tabular}
\vspace{-.5em}
\caption{
\textbf{Noise-level shift} (JiT-B/16, ImageNet 256$\times$256, FID-50K).
We shift the noise level by adjusting $\mu$ in the logit-normal $t$-sampler \cite{esser2024scaling}.
An appropriate noise level is useful, but is not sufficient for addressing the catastrophic failure in $\e$-/$\v$-prediction.
Settings (the same as \cref{tab:3x3s}): 200 epochs, with CFG.
}
\label{tab:mu}
\vspace{-.5em}
\end{table}

\begin{figure}[t]
\centering
\includegraphics[width=0.9\linewidth]{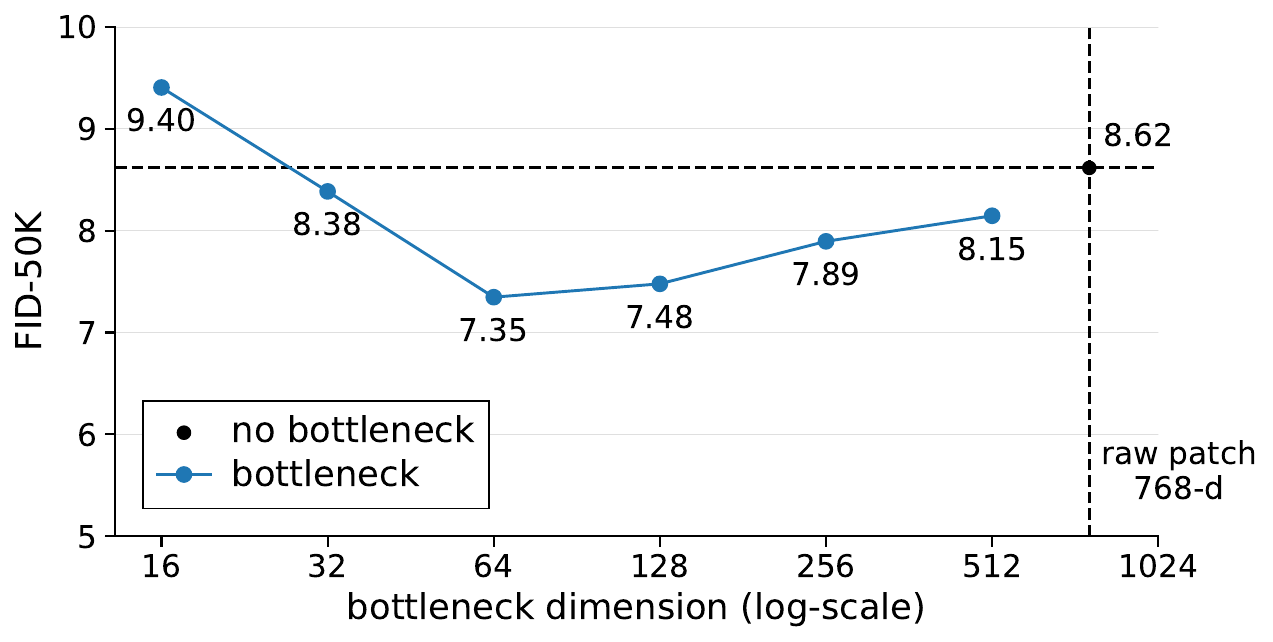}
\vspace{-.7em}
\caption{\textbf{Bottleneck linear embedding.}
Results are for \textbf{JiT-B/16} on ImageNet 256$\times$256. A raw patch is 768-dim ($16{\times}16{\times}3$) and is embedded by two sequential linear layers with an intermediate bottleneck dimension $d'$ ($d' < 768$).  Here, \textbf{\textit{bottleneck embedding is generally beneficial}}, and our $\x$-prediction model can work decently even with aggressive bottlenecks as small as 32 or 16.
Settings (the same as \cref{tab:mu}): 200 epochs, with CFG. 
}
\label{fig:bottleneck}
\vspace{-1em}
\end{figure}

\paragraph{Bottleneck can be beneficial.} Even more surprisingly, we find that, conversely, introducing a bottleneck that \textit{reduces} dimensionality in the network can be beneficial.

Specifically, we turn the linear patch embedding layer into a \textit{low-rank} linear layer, by replacing it with a pair of bottleneck (yet still linear) layers. The first layer reduces the dimension to $d'$, and the second layer expands it to the hidden size of the Transformer. Both layers are linear and serve as a low-rank reparameterization.

\cref{fig:bottleneck} plots the FID \vs the bottleneck dimension $d'$, using JiT-B/16 (768-d per raw patch). Reducing the bottleneck dimension, even to as small as 16-d, does not cause catastrophic failure. In fact, a bottleneck dimension across a wide range (32 to 512) can improve the quality, by a decent margin of up to $\app$1.3 FID. 

From a broader perspective of representation learning, this observation is not entirely unexpected. 
Bottleneck designs are often introduced to encourage the learning of inherently low-dimensional representations \cite{tishby2000information,rifai2011contractive,makhzani2013k,alemi2016deep}.

\definecolor{codeblue}{rgb}{0.25,0.5,0.5}
\definecolor{codekw}{rgb}{0.85, 0.18, 0.50}

\definecolor{codesign}{RGB}{0, 0, 255}
\definecolor{codefunc}{rgb}{0.85, 0.18, 0.50}

\lstdefinelanguage{PythonFuncColor}{
  language=Python,
  keywordstyle=\color{blue}\bfseries,
  commentstyle=\color{codeblue},
  stringstyle=\color{orange},
  showstringspaces=false,
  basicstyle=\ttfamily\small,
  literate=
    {*}{{\color{codesign}* }}{1}
    {-}{{\color{codesign}- }}{1}
    {+}{{\color{codesign}+ }}{1}
    {/}{{\color{codesign}/ }}{1}
    {dataloader}{{\color{codefunc}dataloader}}{1}
    {sample_t}{{\color{codefunc}sample\_t}}{1}
    {randn}{{\color{codefunc}randn}}{1}
    {randn_like}{{\color{codefunc}randn\_like}}{1}
    {jvp}{{\color{codefunc}jvp}}{1}
    {stopgrad}{{\color{codefunc}stopgrad}}{1}
    {l2_loss}{{\color{codefunc}l2\_loss}}{1}
    {net_fn}{{\color{codefunc}net}}{1}
}

\lstset{
  language=PythonFuncColor,
  backgroundcolor=\color{white},
  basicstyle=\fontsize{8pt}{8.4pt}\ttfamily\selectfont,
  columns=fullflexible,
  breaklines=true,
  captionpos=b,
}

\begin{algorithm}[t]
\caption{Training step}
\label{alg:code_train}
\begin{lstlisting}
# net(z, t): JiT network
# x: training batch

t = sample_t()
e = randn_like(x)

z = t * x + (1 - t) * e
v = (x - z) / (1 - t)

x_pred = net_fn(z, t)
v_pred = (x_pred - z) / (1 - t)

loss = l2_loss(v - v_pred)        
\end{lstlisting}
\end{algorithm}
\begin{algorithm}[t]
\caption{Sampling step (Euler)}
\label{alg:code_sample}
\begin{lstlisting}
# z: current samples at t

x_pred = net_fn(z, t)
v_pred = (x_pred - z) / (1 - t)

z_next = z + (t_next - t) * v_pred
\end{lstlisting}
\end{algorithm}

\subsection{Our Algorithm}

Our final algorithm adopts $\x$-prediction and $\v$-loss, which corresponds to \cref{tab:xev}\textbf{(3)(a)}. Formally, we optimize:
\begin{gather}
\mathcal{L} = \mathbb{E}_{t,\x,\e} \Big\|  \v_\theta(\z_t, t) - \v \Big\|^2,
\label{eq:objective}
 \\
\text{where:}\quad \v_\theta(\z_t, t) =  (\net_\theta(\z_t, t)-\bm{z}_t) / (1 - t).
\nonumber
\end{gather}
Alg.\,\ref{alg:code_train} shows the pseudo-code of a training step, and Alg.\,\ref{alg:code_sample} is that of a sampling step (Euler solver; can be extended to Heun or other solvers).
For brevity, class conditioning and CFG \cite{ho2022classifier} are omitted, but both follow standard practice.
To prevent zero division in $1{/}(1{-}t)$ , we clip its denominator (by default, 0.05) whenever computing this division.

\subsection{``Just Advanced'' Transformers}

The strength of a general-purpose Transformer \cite{vaswani2017attention} is partly in that, when its design is decoupled from the specific task, it can benefit from architectural advances developed in other applications. This property underpins the advantage of formulating diffusion with a task-agnostic Transformer.

Following \cite{yao2025reconstruction},
we incorporate popular general-purpose improvements\footnotemark: SwiGLU \cite{shazeer2020glu}, RMSNorm \cite{zhang2019root}, RoPE \cite{su2024roformer}, qk-norm \cite{henry2020query}, all of which were originally developed for language models. 
We also explore in-context class conditioning: but unlike original ViT \cite{Dosovitskiy2021} that appends one class token to the sequence, we appends multiple such tokens (by default, 32; see appendix), following \cite{li2024autoregressive}.
\cref{tab:jat} reports the effects of these components. 

\footnotetext{Our baselines in previous sections all use SwiGLU\,+\,RMSNorm. Removing them has a slight degradation: FID goes from 7.48 to 7.89.}

\section{Comparisons}

\begin{table}[t]
\vspace{-10pt}
\tablestyle{8pt}{1.05}
\begin{tabular}{l|c|c}
 & \textbf{JiT-B/16} & \textbf{JiT-L/16} \\
\shline
{Baseline {(SwiGLU, RMSNorm)}} & 7.48 (6.32) & - \\
\hline
+ RoPE, qk-norm & 6.69 (5.44) & - \\  + {in-context class tokens} & 5.49 (\textbf{4.37}) & 3.39 (\textbf{2.79}) \\
\end{tabular}
\vspace{-.5em}
\caption{\textbf{``Just Advanced'' Transformers}
with \textit{general-purpose} designs.
All are \textbf{JiT/16} for ImageNet 256$\times$256, with bottleneck patch embedding (128-d, \cref{fig:bottleneck}), evaluated by FID-50K. 
Settings: 200 epochs, with CFG (and with CFG interval \cite{kynkaanniemi2024applying} in brackets).
}
\label{tab:jat}
\end{table}

\begin{table}[t]
\centering
\tablestyle{2pt}{1.1}
\begin{tabular}{c | x{32} | ccc | cc | c }
resolution & model & len & patch dim & hiddens & params & Gflops & FID \\
\shline
256$\times$256 & JiT-B/{16} & 256 & 768 & 768 & 131 & 25 & 4.37 \\
512$\times$512 & JiT-B/{32} & 256 & 3072 & 768 & 133 & 26 & 4.64 \\
\textbf{1024$\times$1024} & JiT-B/\textbf{64} & 256 & 12288 & 768 & 141 & 30 & 4.82 \\ 
\end{tabular}
\vspace{-.5em}
\caption{\textbf{ImageNet 1024$\times$1024 with JiT-B/64}.
All entries have roughly the same number of parameters and compute.
Settings: if not specified here, the same as \cref{tab:jat} (all are with CFG interval).
}
\label{tab:in1024}
\end{table}

\paragraph{High-resolution generation on pixels.} 
In \cref{tab:in1024}, we further report our \textit{base}-size model (JiT-B) on ImageNet at resolutions {512} and even {1024}.
We use patch sizes \textit{proportional} to image sizes, and therefore the sequence length at different resolutions remains the same.
The per-patch dimension can be as high as 3072 or 12288, and \textit{none} of the common models would have sufficient hidden units.

\newcommand{\deemph}[1]{\textcolor{black!30}{#1}}

\begin{table}[t]
\centering
\tablestyle{2pt}{1.1}
\begin{tabular}{c | x{32}x{32} }
\textbf{256$\times$256} & \deemph{200-ep} & 600-ep \\
\shline
{JiT-B/\textbf{16}} & \deemph{4.37} & 3.66 \\
{JiT-L/\textbf{16}} & \deemph{2.79} & 2.36 \\
{JiT-H/\textbf{16}} & \deemph{2.29} & 1.86   \\
{JiT-G/\textbf{16}} & \deemph{2.15} & {\textbf{1.82}}   \\
\end{tabular}
\hspace{1em}
\tablestyle{2pt}{1.1}
\begin{tabular}{c | x{32}x{32} }
\textbf{512$\times$512} & \deemph{200-ep} & 600-ep \\
\shline
{JiT-B/\textbf{32}}  & \deemph{4.64} & 4.02 \\
{JiT-L/\textbf{32}} & \deemph{3.06} & 2.53 \\
 {JiT-H/\textbf{32}} & \deemph{2.51} & {1.94} \\
{JiT-G/\textbf{32}} & \deemph{2.11} & \textbf{1.78}   \\
\end{tabular}
\vspace{-.5em}
\caption{\textbf{Scalability on ImageNet 256$\times$256 and 512$\times$512}, evaluated by FID-50K.
All models have the \textit{same} sequence length of 16$\times$16, and thus the models at 512 resolution have nearly the same compute as their 256 counterparts.
Settings: the same as \cref{tab:in1024}.
}
\label{tab:scale}
\vspace{-.8em}
\end{table}

\cref{tab:in1024} shows that our models perform decently across resolutions. All models have similar numbers of parameters and computational cost, which only differ in the input/output patch embeddings. Our method does not suffer from the curse of observed dimensionalities.

\paragraph{Scalability.} A core goal of \textit{decoupling} the Transformer design with the task is to leverage the potential for scalability.
\cref{tab:scale} provides results on ImageNet 256 and 512 with four model sizes (note that at resolution 512, none of these models have more hidden units than the patch dimension). The model sizes and flops are shown in \cref{tab:in256-sys} and \ref{tab:in512-sys}: our model at resolution 256 has similar cost as its counterpart at 512. Our approach benefits from scaling.

Interestingly, the FID difference between resolution 256 and 512 becomes smaller with larger models. For JiT-G, the FID at 512 resolution is even lower. For very large models on ImageNet, FID performance largely depends on overfitting, and denoising at 512 resolution poses a more challenging task, making it less susceptible to overfitting. 

\paragraph{Reference results from previous works.} As a reference, we compare with previous results in \cref{tab:in256-sys} and \ref{tab:in512-sys}. We mark the \textit{pre-training} components involved for each method.
Compared with other pixel-based methods, ours is purely driven by \textit{plain}, general-purpose Transformers. Our models are \textit{compute-friendly} and avoid the \textit{quadratic} scaling of expense with doubled resolution  (see flops in \cref{tab:in512-sys}).

Our approach does not use extra losses or pre-training, which may lead to further gains (an example is in appendix). These directions are left for future work.

\newcommand{\viswidth}{0.195\linewidth}
\begin{figure}[t]
\vspace{0.75em}
\centering
\includegraphics[width=\viswidth]{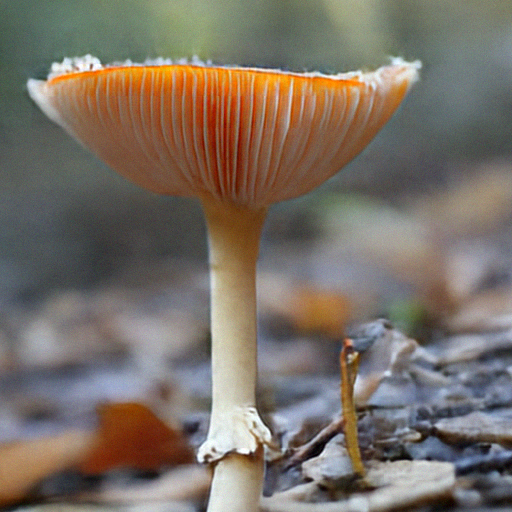}\hhs
\includegraphics[width=\viswidth]{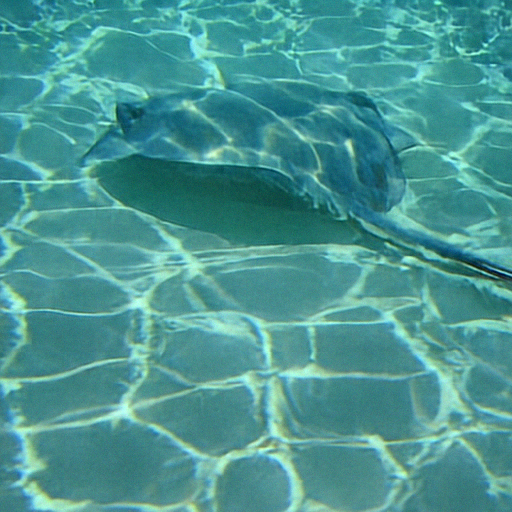}\hhs
\includegraphics[width=\viswidth]{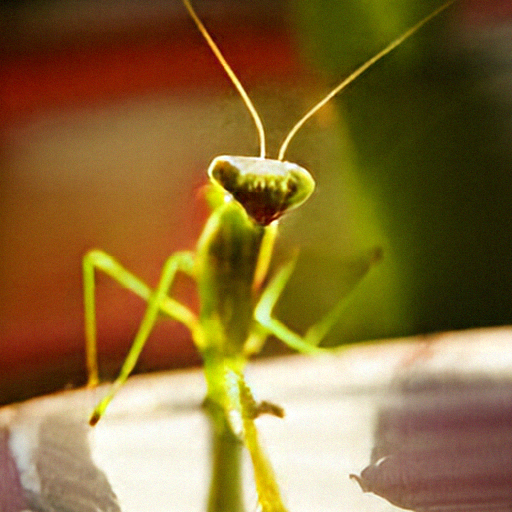}\hhs
\includegraphics[width=\viswidth]{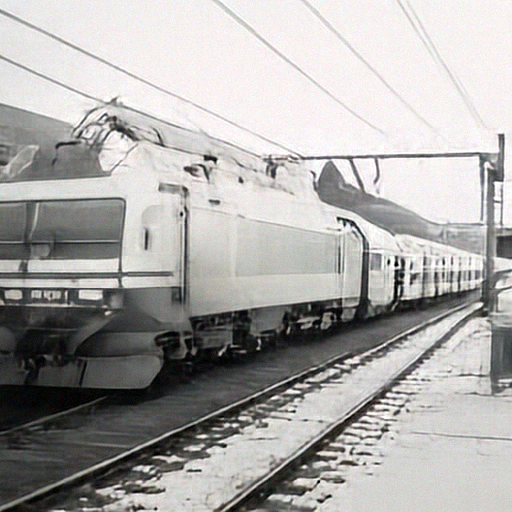}\hhs
\includegraphics[width=\viswidth]{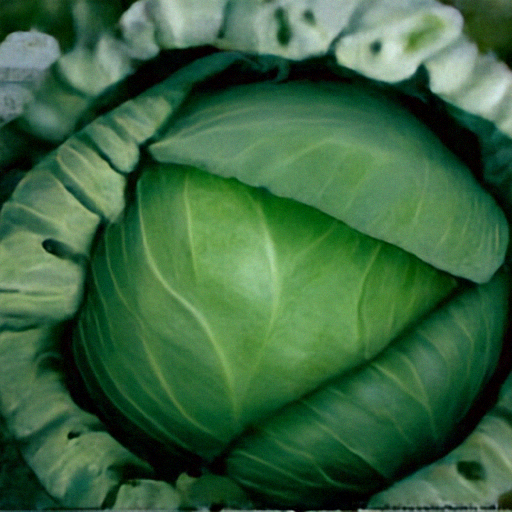}\vvs
\\
\includegraphics[width=\viswidth]{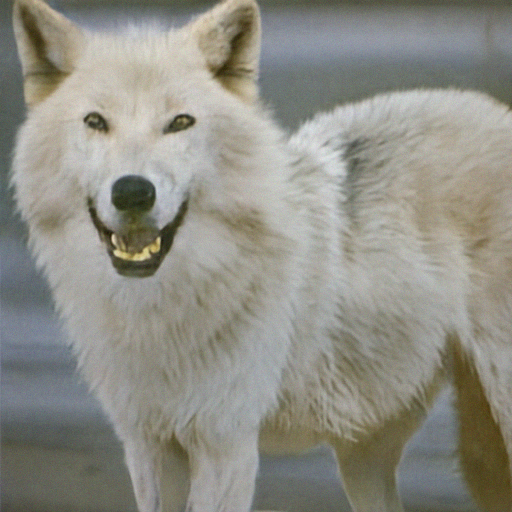}\hhs
\includegraphics[width=\viswidth]{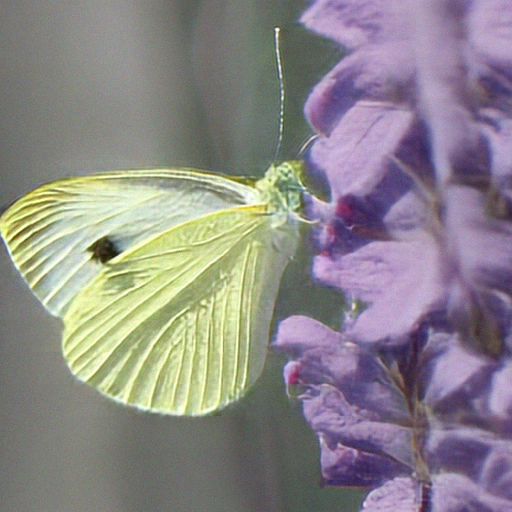}\hhs
\includegraphics[width=\viswidth]{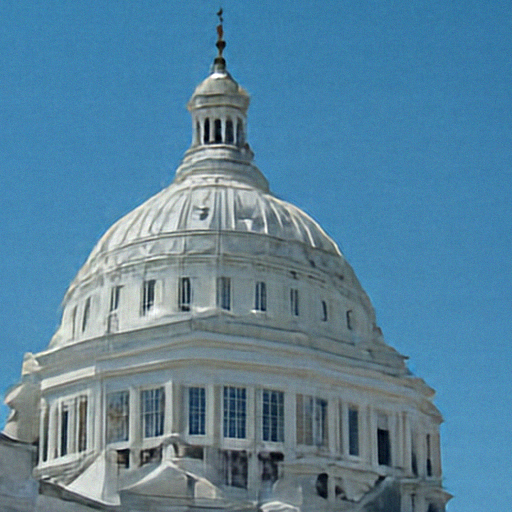}\hhs
\includegraphics[width=\viswidth]{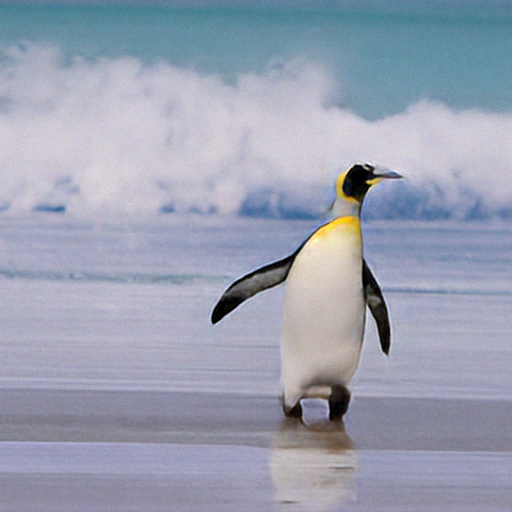}\hhs
\includegraphics[width=\viswidth]{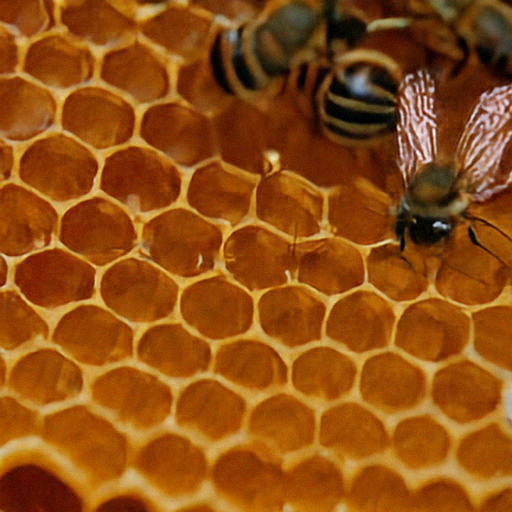}\vvs
\\
\includegraphics[width=\viswidth]{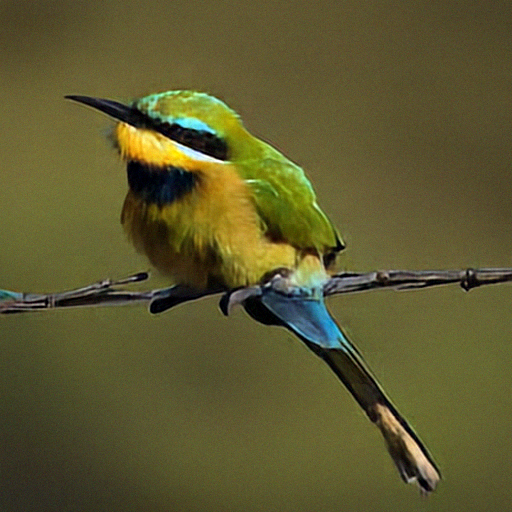}\hhs
\includegraphics[width=\viswidth]{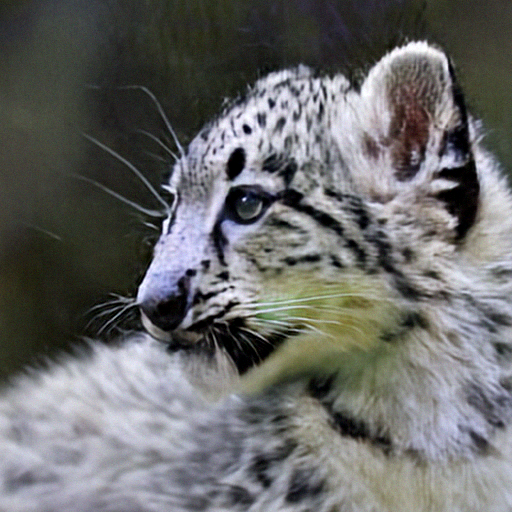}\hhs
\includegraphics[width=\viswidth]{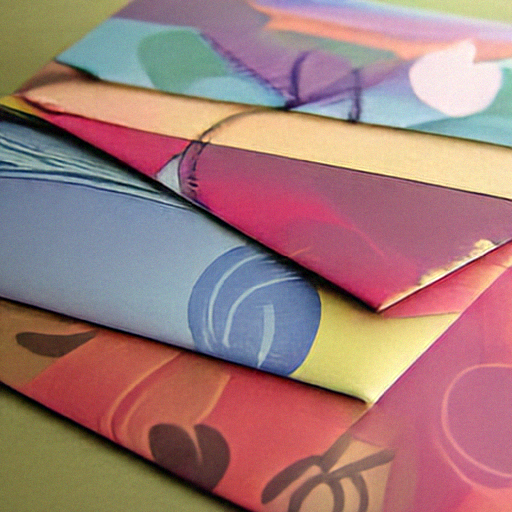}\hhs
\includegraphics[width=\viswidth]{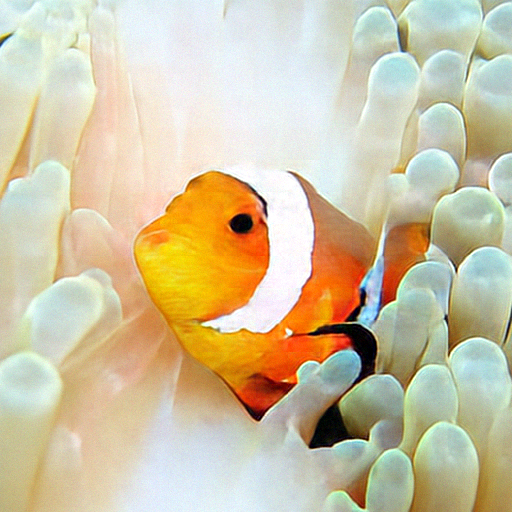}\hhs
\includegraphics[width=\viswidth]{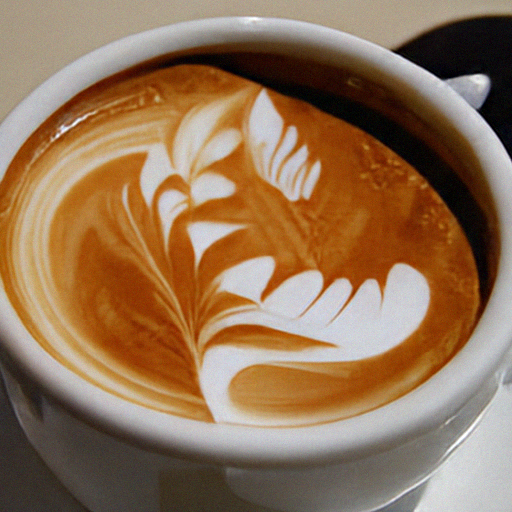}\vvs
\\
\includegraphics[width=\viswidth]{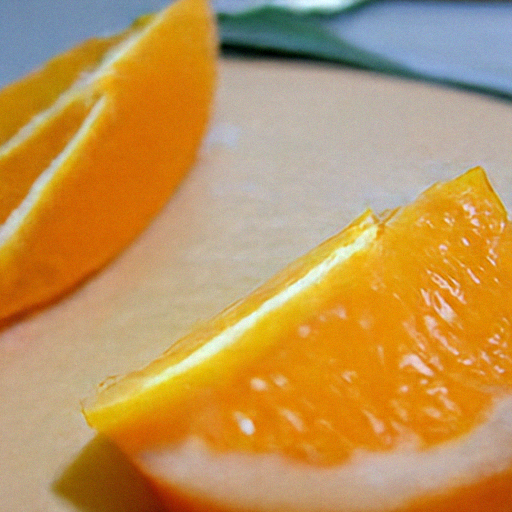}\hhs
\includegraphics[width=\viswidth]{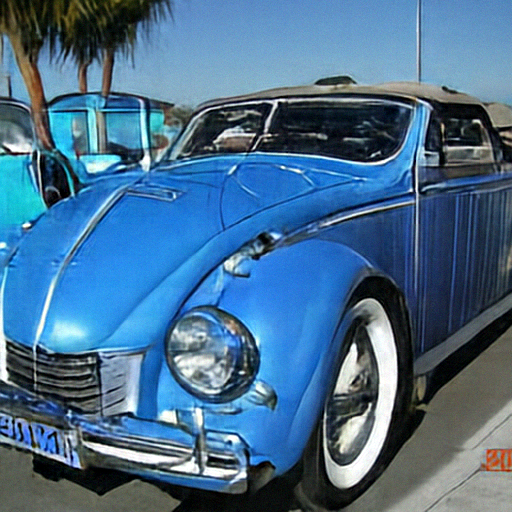}\hhs
\includegraphics[width=\viswidth]{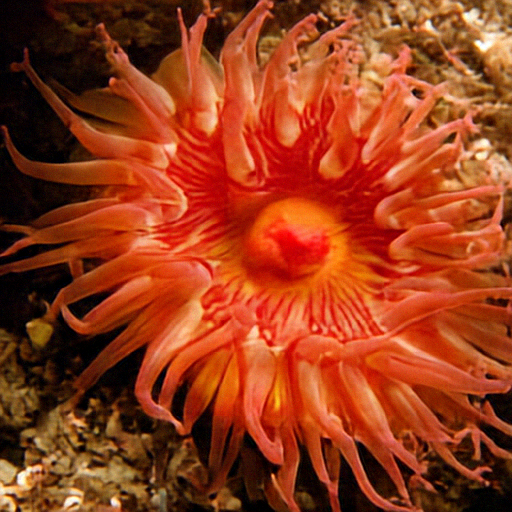}\hhs
\includegraphics[width=\viswidth]{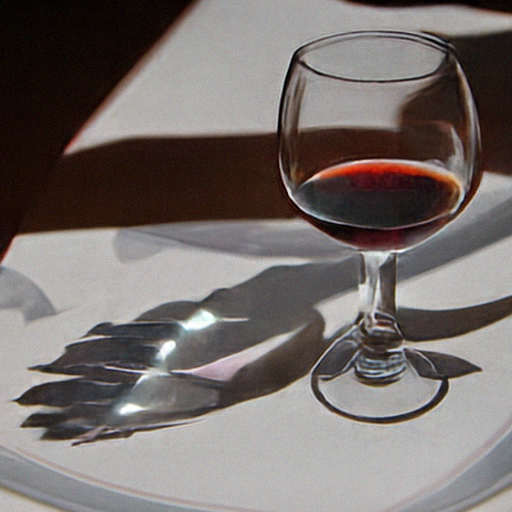}\hhs
\includegraphics[width=\viswidth]{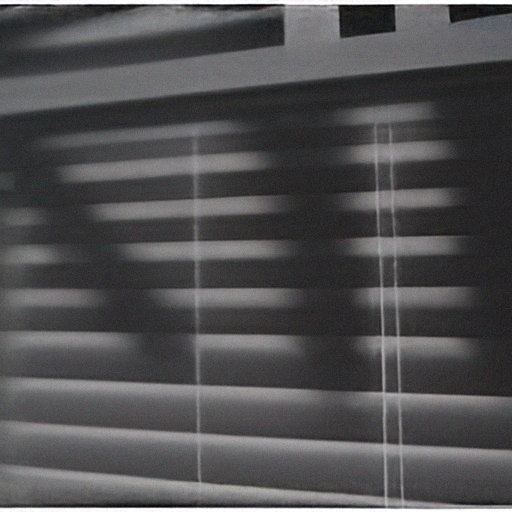}\vvs
\\
\includegraphics[width=\viswidth]{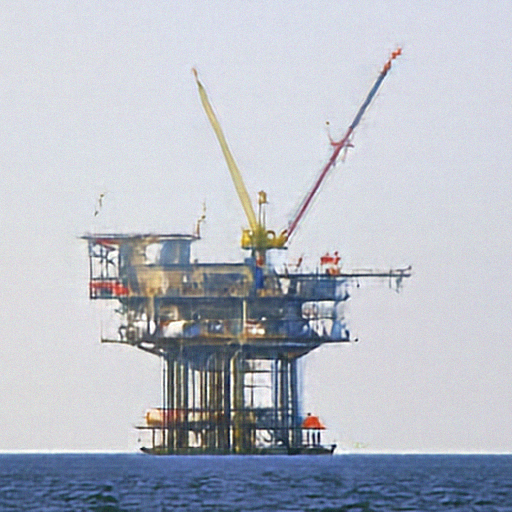}\hhs
\includegraphics[width=\viswidth]{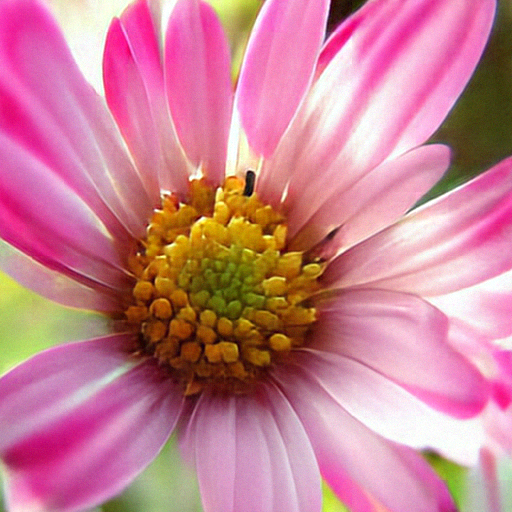}\hhs
\includegraphics[width=\viswidth]{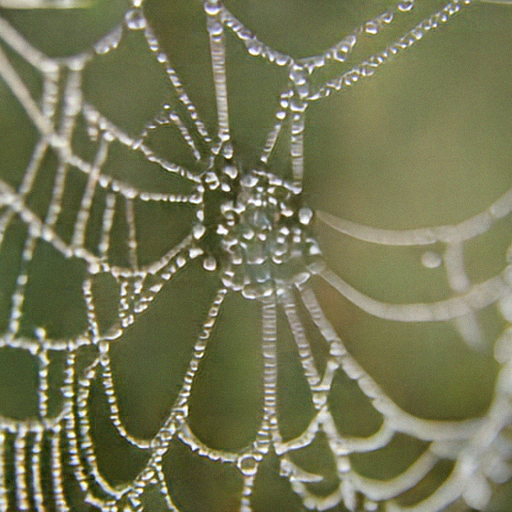}\hhs
\includegraphics[width=\viswidth]{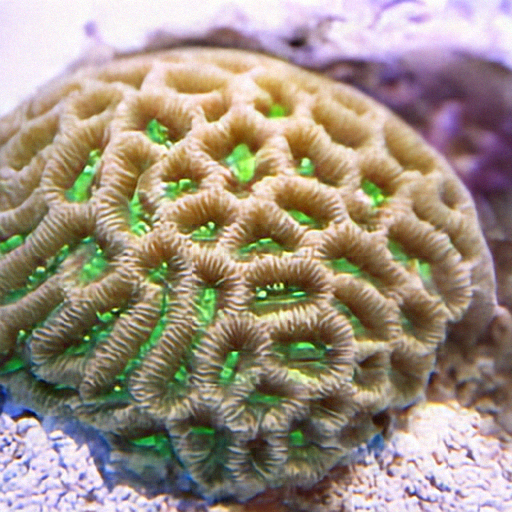}\hhs
\includegraphics[width=\viswidth]{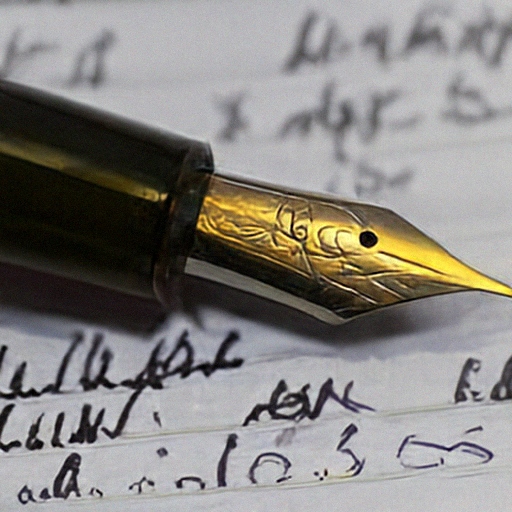}\vvs
\\
\includegraphics[width=\viswidth]{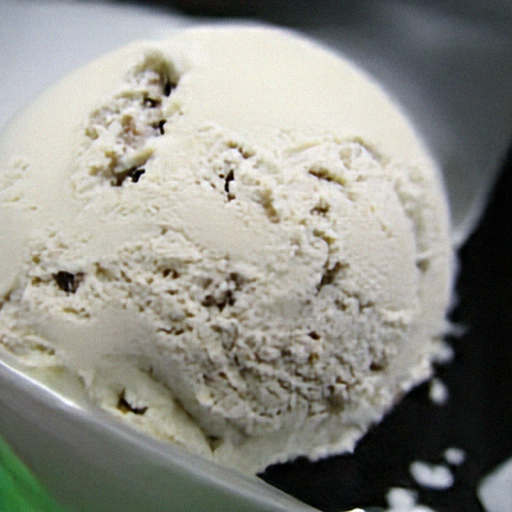}\hhs
\includegraphics[width=\viswidth]{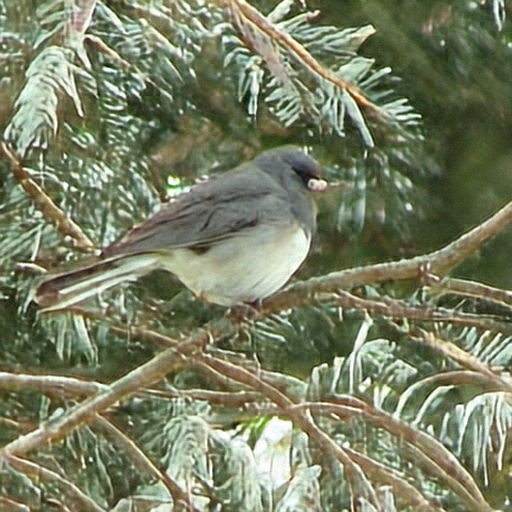}\hhs
\includegraphics[width=\viswidth]{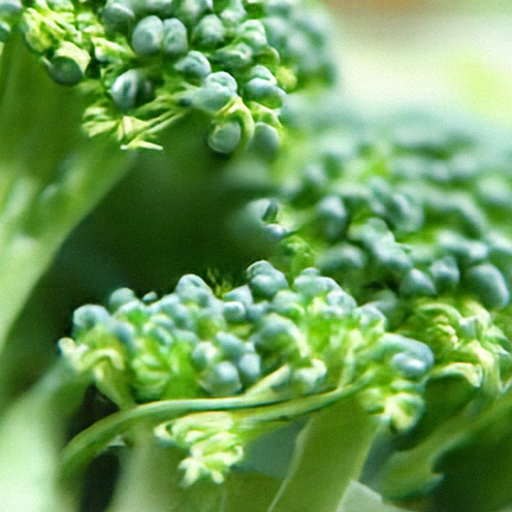}\hhs
\includegraphics[width=\viswidth]{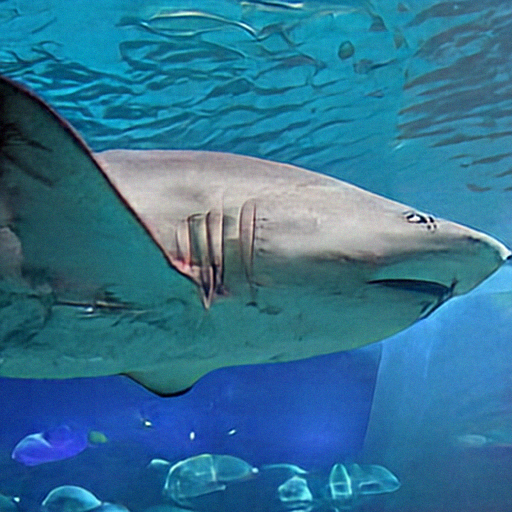}\hhs
\includegraphics[width=\viswidth]{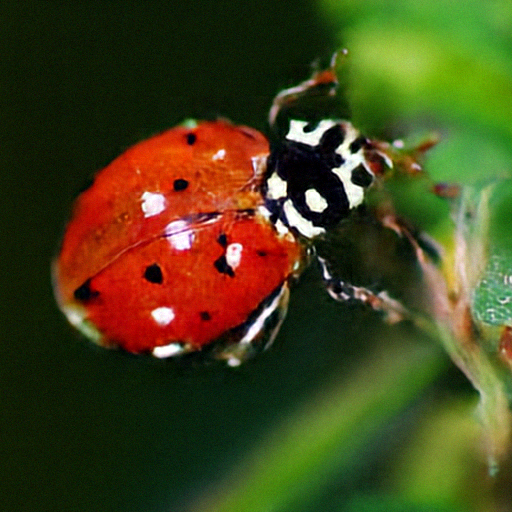}\vvs
\caption{\textbf{Qualitative Results.} Selected examples on ImageNet 512$\times$512 using JiT-H/32. More uncurated results are in appendix.
}
\label{fig:results}
\end{figure}

\newcommand\headspace{\hspace{.2em}}
\newcommand\shrink[1]{{\fontsize{6pt}{7.2pt}\selectfont{#1}}}

\begin{table}[t]
\tablestyle{1pt}{1.1}
\scriptsize
\begin{tabular}{l | c c c | c c | c | c}
 & \multicolumn{3}{c|}{\textbf{pre-training}} & & & & \\
\textbf{ImgNet 256$\times$256} & token & perc. & self-sup. &
 \multirow[t]{2}{*}{\scriptsize\shortstack{\textbf{params}}} &
 \multirow[t]{2}{*}{\scriptsize\shortstack{\textbf{Gflops}}} &
 {\textbf{FID}$\downarrow$} & {\textbf{IS}$\uparrow$} \\
\shline
\rowcolor[gray]{0.9} \multicolumn{8}{l}{\textit{Latent-space Diffusion}}  \\
\headspace DiT-XL/2 \cite{Peebles2023} & \shrink{SD-VAE} & \shrink{VGG} & - & 675+49M & 119 & 2.27 & 278.2 \\
\headspace SiT-XL/2 \cite{ma2024sit} & \shrink{SD-VAE} & \shrink{VGG} & - & 675+49M & 119 & 2.06 & 277.5 \\
\headspace REPA \cite{repa}, SiT-XL/2  & \shrink{SD-VAE} & \shrink{VGG} & \shrink{DINOv2} & 675+49M & 119 & 1.42 & 305.7 \\
\headspace 
\shrink{LightningDiT-XL/2} \cite{yao2025reconstruction} & \shrink{VA-VAE} & \shrink{VGG} & \shrink{DINOv2} & 675+49M & 119 & 1.35 & 295.3 \\
\headspace DDT-XL/2 \cite{wang2025ddt} & \shrink{SD-VAE} & \shrink{VGG} & \shrink{DINOv2} & 675+49M & 119 & 1.26 & 310.6  \\
\headspace RAE \cite{zheng2025diffusion}, DiT$^{\text{DH}}$-XL/2 & \shrink{RAE} & \shrink{VGG} & \shrink{DINOv2} & \hspace{0pt} 839+415M & 146 & \textbf{1.13} & 262.6 \\
\midline
\rowcolor[gray]{0.9} \multicolumn{8}{l}{\textit{\textbf{Pixel}-space (non-diffusion)}}  \\
\headspace JetFormer \cite{tschannen2024jetformer} & - & - & - & 2.8B & - & 6.64 & - \\
\headspace FractalMAR-H \cite{li2025fractal} & - & - & - & 848M & - & 6.15 & 348.9 \\
\midline
\rowcolor[gray]{0.9} \multicolumn{8}{l}{\textit{\textbf{Pixel}-space Diffusion}}  \\
\headspace ADM-G \cite{dhariwal2021diffusion} & - & - & - & 554M & 1120 & 4.59 & 186.7 \\
\headspace RIN \cite{jabri2022scalable} & - & - & - & 410M & 334 & 3.42 & 182.0 \\
\headspace SiD \cite{hoogeboom2023simple} , UViT/2  & - & - & - & 2B & 555 & 2.44 & 256.3 \\
\headspace VDM++, UViT/2 & - & - & - & 2B & 555 & 2.12 &  267.7 \\
\headspace SiD2 \cite{hoogeboom2024simpler}, UViT/2 & - & - & - & N/A & 137 & 1.73 & - \\
\headspace SiD2 \cite{hoogeboom2024simpler}, UViT/1 & - & - & - & N/A & 653 & \textbf{1.38} & - \\
\headspace PixelFlow \cite{chen2025pixelflow}, XL/4 & - & - & - & 677M & 2909 & 1.98 & 282.1 \\
\headspace PixNerd \cite{wang2025pixnerd}, XL/16 & - & - &  \shrink{DINOv2} & 700M & 134 & 2.15 & 297 \\
\cline{1-8}
\headspace \textbf{JiT-B/16} & - & - & - & 131M & 25 & 3.66 & 275.1 \\
\headspace \textbf{JiT-L/16} & - & - & - & 459M & 88 & 2.36 & 298.5 \\
\headspace \textbf{JiT-H/16} & - & - & - & 953M & 182 & 1.86 & 303.4 \\
\headspace \textbf{JiT-G/16} & - & - & - & 2B & 383 & 1.82 & 292.6
\\
\end{tabular}
\vspace{-.2em}
\caption{
\textbf{Reference results on ImageNet 256$\times$256.} 
FID \cite{heusel2017gans} and IS \cite{salimans2016improved} of 50K samples are evaluated. The ``pre-training'' columns list the external models required to obtain the results (note that the perceptual loss \cite{zhang2018unreasonable} uses a pre-trained VGG classifier \cite{simonyan2014very}).
The parameters include the generator and tokenizer decoder (used at inference-time), but exclude other pre-trained components.
The Giga-flops are measured for a single forward pass (not counting the tokenizer) and are roughly proportional to the computational cost of an iteration during both training and inference (for the multi-scale method \cite{chen2025pixelflow}, we measure the finest level).
\label{tab:in256-sys}
\vspace{2em}
}

\tablestyle{1pt}{1.1}
\scriptsize
\begin{tabular}{l | c c c | c c | c | c}
 & \multicolumn{3}{c|}{\textbf{pre-training}} & & & & \\
\textbf{ImgNet 512$\times$512} & token & perc. & self-sup. &
 \multirow[t]{2}{*}{\scriptsize\shortstack{\textbf{params}}} &
 \multirow[t]{2}{*}{\scriptsize\shortstack{\textbf{Gflops}}} &
 {\textbf{FID}$\downarrow$} & {\textbf{IS}$\uparrow$} \\
\shline
\rowcolor[gray]{0.9} \multicolumn{8}{l}{\textit{Latent-space Diffusion}}  \\
\headspace DiT-XL/2 \cite{Peebles2023} & \shrink{SD-VAE} & \shrink{VGG} & - & 675+49M & 525 & 3.04 & 240.8 \\
\headspace SiT-XL/2 \cite{ma2024sit} & \shrink{SD-VAE} & \shrink{VGG} & - & 675+49M & 525 & 2.62 & 252.2 \\
\headspace REPA \cite{repa}, SiT-XL/2  & \shrink{SD-VAE} & \shrink{VGG} & \shrink{DINOv2} & 675+49M & 525 & 2.08 & 274.6 \\
\headspace DDT-XL/2 \cite{wang2025ddt} & \shrink{SD-VAE} & \shrink{VGG} & \shrink{DINOv2} & 675+49M & 525 & 1.28 & 305.1 \\
\headspace RAE \cite{zheng2025diffusion}, DiT$^{\text{DH}}$-XL/2 & \shrink{RAE} & \shrink{VGG} & \shrink{DINOv2} & \hspace{0pt} 839+415M & 642 & \textbf{1.13} & 259.6 \\
\midline
\rowcolor[gray]{0.9} \multicolumn{8}{l}{\textit{\textbf{Pixel}-space Diffusion}}  \\
\headspace ADM-G \cite{dhariwal2021diffusion} & - & - & - & 559M & 1983 & 7.72 & 172.7 \\
\headspace RIN \cite{jabri2022scalable} & - & - & - & 320M & 415 & 3.95 & 216.0 \\
\headspace SiD \cite{hoogeboom2023simple} , UViT/4  & - & - & - & 2B & 555 & 3.02 & 248.7 \\
\headspace VDM++, UViT/4 & - & - & - & 2B & 555 & 2.65 & 278.1\\
\headspace SiD2 \cite{hoogeboom2024simpler}, UViT/4 & - & - & - & N/A & 137 & 2.19 & - \\
\headspace SiD2 \cite{hoogeboom2024simpler}, UViT/2 & - & - & - & N/A & 653 & \textbf{1.48} & - \\
\headspace PixNerd \cite{wang2025pixnerd}, XL/16 & - & - &  \shrink{DINOv2} & 700 M & 583 & 2.84 & 245.6 \\
\cline{1-8}
\headspace \textbf{JiT-B/32} & - & - & - & 133M & 26 & 4.02 & 271.0 \\
\headspace \textbf{JiT-L/32} & - & - & - & 462M & 89 & 2.53 & 299.9 \\
\headspace \textbf{JiT-H/32} & - & - & - & 956M & 183 & 1.94 & 309.1 \\
\headspace \textbf{JiT-G/32} & - & - & - & 2B & 384 & 1.78 & 306.8
\end{tabular}
\vspace{-.2em}
\caption{
\textbf{Reference results on ImageNet 512$\times$512.} JiT has an \textit{aggressive} patch size and can use \textit{small} compute to achieve strong results. Notations are similar to \cref{tab:in256-sys}.
}
\label{tab:in512-sys}
\vspace{-.5em}
\end{table}

\section{Discussion and Conclusion}

\textit{Noise is inherently different from natural data}. Over the years, the development of diffusion models has focused primarily on probabilistic formulations, while paying less attention to the capabilities (and limitations) of the neural networks used. However, neural networks are not infinitely capable, and they can better use their capacity to model data rather than noise. Under these perspectives, our findings on $\x$-prediction are, in hindsight, a natural outcome.

Our work adopts a minimalist and self-contained design. By reducing domain-specific inductive biases, we hope our approach can generalize to other domains where tokenizers are difficult to obtain.
This property is particularly desirable for scientific applications that involve raw, high-dimensional natural data.
We envision that the general-purpose ``Diffusion\,+\,Transformer'' paradigm will be a potential foundation in other areas.

\newpage

\appendix
\section{Implementation Details}
Our implementation closely follows the public codebases of DiT \cite{Peebles2023} and SiT \cite{ma2024sit}.
Our configurations are summarized in \cref{tab:config}.
We describe the details as follows.

\paragraph{Time distribution.} Following \cite{esser2024scaling}, during training, we adopt the logit-normal distribution over $t$ \cite{esser2024scaling}: $\text{logit}(t) \sim \mathcal{N}(\mu,\sigma^2)$.
Specifically, we sample $s\sim \mathcal{N}(\mu,\sigma^2)$ and let $t{=}\text{sigmoid}(s)$. The hyper-parameter $\mu$ determines the noise level (see \cref{tab:mu}), and by default we set $\mu$ = --0.8 on ImageNet at resolution 256 (or 512, 1024), and fix $\sigma$ as 0.8.

\paragraph{ImageNet 512${\times}$512 and 1024${\times}$1024.} We adopt \textbf{JiT/32} (\ie, a patch size of 32) on ImageNet 512${\times}$512. The model leads to a sequence of 256 = 16$\times$16 patches, the same as JiT/16 on ImageNet 256${\times}$256. As such, JiT/32 only differs from JiT/16 in the input/output patch dimension, increasing from 768-d to 3072-d per patch; all other computations and costs are exactly the same.

To reuse the exact same recipe from ImageNet 256${\times}$256, for 512${\times}$512 images we rescale the magnitude of the noise $\e$ by 2$\times$: that is, $\e \sim \mathcal{N}(0, 2^2 \mathbf{I})$. This simple modification approximately maintains the signal-to-noise ratio (SNR) between the 256${\times}$256 and 512${\times}$512 resolutions \cite{hoogeboom2023simple,chen2023importance,hoogeboom2024simpler}.
No other changes to the ImageNet 256${\times}$256 configuration are required or applied.

For ImageNet 1024${\times}$1024, we use the model JiT/64 and scale the noise $\e$ by 4$\times$. No other change is needed.

\paragraph{In-context class conditioning.} Standard DiT \cite{Peebles2023} performs class conditioning through adaLN-Zero.
In \cref{tab:jat}, we further explore in-context class-conditioning.

Specifically, following ViT \cite{Dosovitskiy2021}, one can prepend a class token to the sequence of patches. This is referred to as ``in-context conditioning'' in DiT \cite{Peebles2023}. 
Note that we use in-context conditioning jointly with the default adaLN-Zero conditioning, unlike DiT.
In addition, following MAR \cite{li2024autoregressive}, we further prepend multiple such tokens to the sequence. These tokens are repeated instances of the same class token, with different positional embeddings added. 
We prepend 32 such tokens.
Moreover, rather than prepending these tokens to the Transformer's input, we find that prepending them at later blocks can be beneficial (see ``in-context start block'' in \cref{tab:config}). 
\cref{tab:jat} shows that our implementation of in-context conditioning improves FID by $\app$1.2.

\begin{table}[t]
\centering
\resizebox{1.0\width}{!}{
\tablestyle{4pt}{1.02}
\begin{tabular}{l | x{24}x{24}x{24}x{24}}
 & \textbf{JiT-B} & \textbf{JiT-L} & \textbf{JiT-H} & \textbf{JiT-G} \\
\shline
\rowcolor[gray]{0.9}\multicolumn{5}{l}{\textbf{architecture}} \\
depth & 12 & 24 & 32 & 40 \\
hidden dim & 768 & 1024 & 1280 & 1664 \\
heads & 12 & 16 & 16 & 16 \\
image size & \multicolumn{4}{c}{256 (other settings: 512, or 1024)} \\
patch size & \multicolumn{4}{c}{\texttt{image\_size} / 16} \\
bottleneck & \multicolumn{4}{c}{128 (B/L), 256 (H/G)} \\
dropout & \multicolumn{4}{c}{{0 (B/L), {0.2} (H/G)}} \\
in-context class tokens & \multicolumn{4}{c}{32 (if used)} \\
in-context start block & 4 & 8 & 10 & 10 \\
\midline
\rowcolor[gray]{0.9}\multicolumn{5}{l}{\textbf{training}} \\
epochs & \multicolumn{4}{c}{{200 (ablation), 600}} \\
warmup epochs \cite{Goyal2017} & \multicolumn{4}{c}{5} \\
optimizer & \multicolumn{4}{c}{Adam \cite{adam2014method}, $\beta_1, \beta_2=0.9, 0.95$} \\
batch size & \multicolumn{4}{c}{1024} \\
learning rate & \multicolumn{4}{c}{2e-4} \\ 
learning rate schedule & \multicolumn{4}{c}{constant} \\
weight decay & \multicolumn{4}{c}{0} \\ 
ema decay & \multicolumn{4}{c}{\{0.9996, 0.9998, 0.9999\}} \\
time sampler & \multicolumn{4}{c}{$\text{logit}(t){\sim}\mathcal{N}(\mu, \sigma^2)$, $\mu$ = --0.8, $\sigma$ = 0.8 } \\
noise scale & \multicolumn{4}{c}{{1.0 $\times$ \texttt{image\_size} / 256}} \\
clip of $(1-t)$ in division & \multicolumn{4}{c}{0.05} \\
class token drop (for CFG) & \multicolumn{4}{c}{0.1} \\
\midline
\rowcolor[gray]{0.9}\multicolumn{5}{l}{\textbf{sampling}} \\
ODE solver & \multicolumn{4}{c}{Heun \cite{heun1900neue}} \\
ODE steps & \multicolumn{4}{c}{50} \\
time steps & \multicolumn{4}{c}{linear in [0.0, 1.0]} \\
CFG scale sweep range \cite{ho2022classifier} & \multicolumn{4}{c}{[1.0, 4.0]}  \\
CFG interval \cite{kynkaanniemi2024applying} & \multicolumn{4}{c}{{[0.1, 1] (if used)}} \\
\end{tabular}
}
\vspace{-.5em}
\caption{\textbf{Configurations of experiments.}}
\vspace{-.5em}
\label{tab:config}
\end{table}

\begin{figure}[t]
\centering
\includegraphics[width=0.8\linewidth]{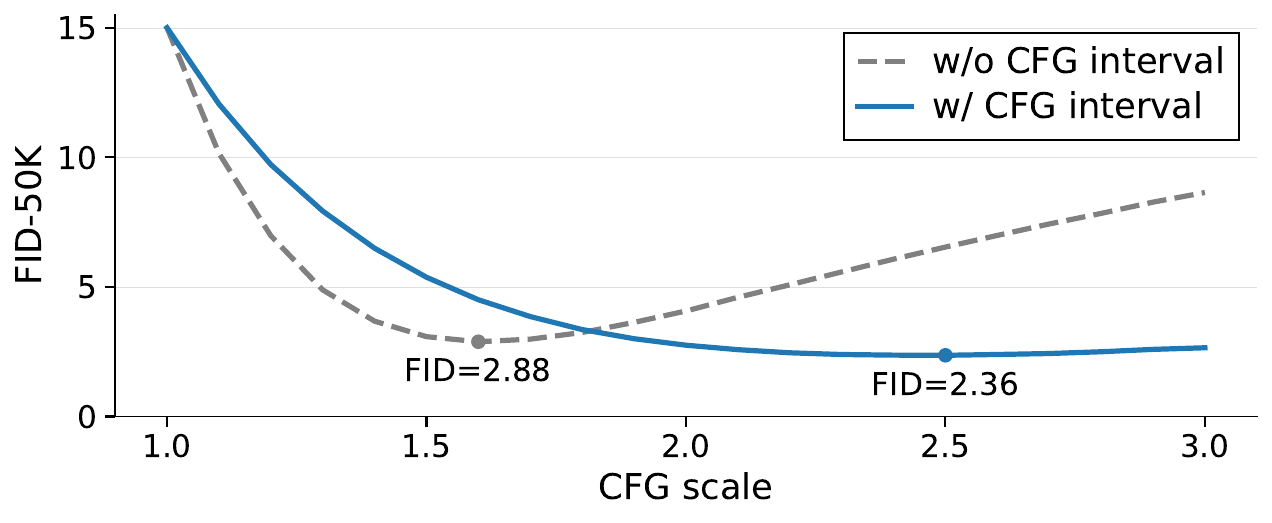}
\vspace{-.7em}
\caption{The influence of CFG, without and with CFG interval (for JiT-L/16 on ImageNet 256$\times$256, 600 epochs).
}
\label{fig:cfg}
\end{figure}

\paragraph{Dropout and early stop.} We apply dropout \cite{srivastava2014dropout} in JiT-H and G to mitigate the risk of overfitting.
Specifically, we apply dropout to the middle half of the Transformer blocks.
For Transformer blocks with dropout, we apply it to both the attention block and the MLP block.

As the G-size models still tend to overfit under our current dropout setting, we apply early stopping when the monitored FID begins to degrade. This occurs at around 320 epochs for both JiT-G/16 and JiT-G/32.

\paragraph{EMA and CFG.} Our study covers a wide range of configurations, including variations in loss and prediction spaces, model sizes, and architectural components.
The optimal values of the CFG scale \cite{ho2022classifier} and EMA (exponential moving average) decay vary from case to case, and fixing them may lead to incomplete or misleading observations. Since maintaining these hyperparameter configurations is relatively inexpensive, we strive to adopt the optimal values.

Specifically, for the CFG scale $\omega$ \cite{ho2022classifier}, we determine the optimal value by searching over a range of candidate scales at inference time, as is common practice in existing work. For EMA decays, we maintain multiple copies of the moving-averaged parameters during training, which introduces a negligible computational overhead. To avoid memory overhead, different EMA copies can be stored on separate devices (\eg, GPUs). As such, both the CFG scale and EMA decay are essentially inference-time decisions.

Our CFG scale candidates range from 1.0 to 4.0 with a step size of 0.1. 
The influence of CFG is shown in \cref{fig:cfg} for a JiT-L/16 model.
Our EMA decay candidates are 0.9996, 0.9998, and 0.9999, evaluated with a batch size of 1024. 
For each model (including any one in the ablation), we search for the optimal setting using 8K samples and then apply it to evaluate 50K samples.

\paragraph{Evaluation.} Following common pratice, we evaluate the FID \cite{heusel2017gans} against the ImageNet training set. We evaluate FID on 50K generated images, with 50 samples for each of the 1000 ImageNet classes. We evaluate the Inception Score (IS) \cite{salimans2016improved} on the same 50K images.

\section{Additional Experiments}

\begin{figure}[t]
\centering
\includegraphics[width=0.8\linewidth]{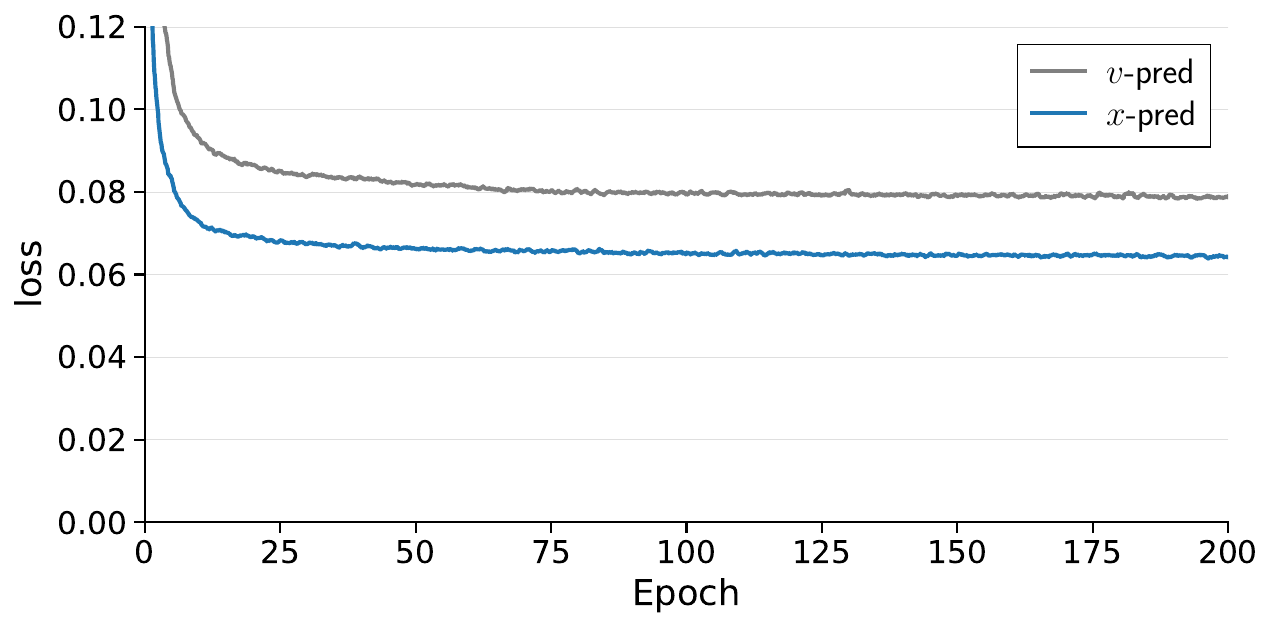}
\tablestyle{.5pt}{1.0}
\begin{tabular}{@{}>{\centering\arraybackslash}m{0.22\linewidth}
                >{\centering\arraybackslash}m{0.15\linewidth}
                >{\centering\arraybackslash}m{0.15\linewidth}
                >{\centering\arraybackslash}m{0.15\linewidth}
                >{\centering\arraybackslash}m{0.15\linewidth}
                >{\centering\arraybackslash}m{0.15\linewidth}@{}}
&
\scriptsize{$t=0.1$} &
\scriptsize{$t=0.2$} &
\scriptsize{$t=0.3$} &
\scriptsize{$t=0.4$} &
\scriptsize{$t=0.5$}
\\
\scriptsize{noisy image} &
\includegraphics[width=1.0\linewidth]{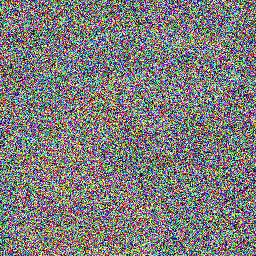} &
\includegraphics[width=1.0\linewidth]{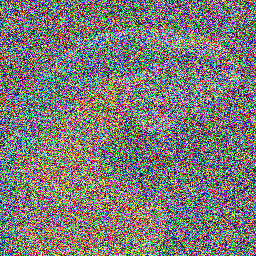} &
\includegraphics[width=1.0\linewidth]{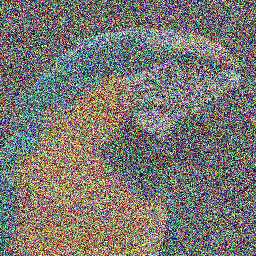} &
\includegraphics[width=1.0\linewidth]{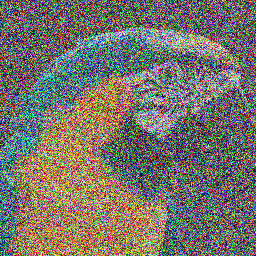} &
\includegraphics[width=1.0\linewidth]{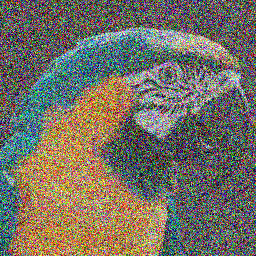}
\\
{\scriptsize{denoised image from
\textbf{$\x$-pred}}} &
\includegraphics[width=1.0\linewidth]{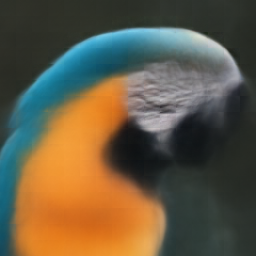} &
\includegraphics[width=1.0\linewidth]{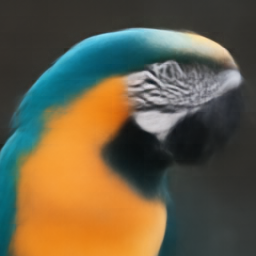} &
\includegraphics[width=1.0\linewidth]{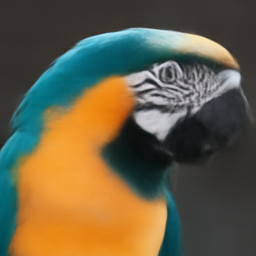} &
\includegraphics[width=1.0\linewidth]{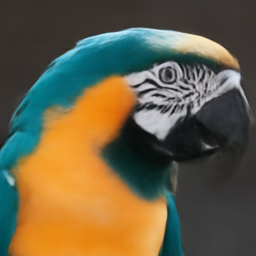} &
\includegraphics[width=1.0\linewidth]{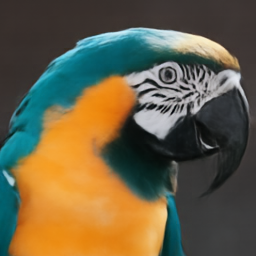}
\\
{\scriptsize{denoised image from
\textbf{$\v$-pred}}} &
\includegraphics[width=1.0\linewidth]{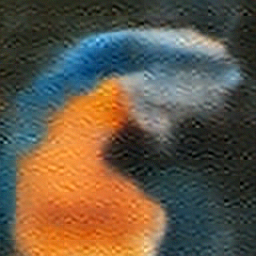} &
\includegraphics[width=1.0\linewidth]{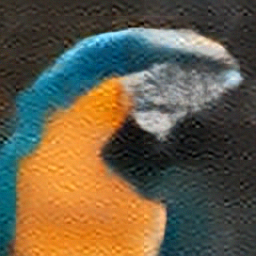} &
\includegraphics[width=1.0\linewidth]{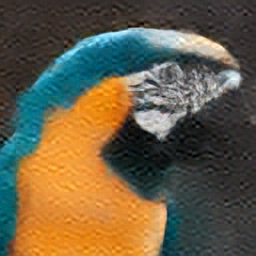} &
\includegraphics[width=1.0\linewidth]{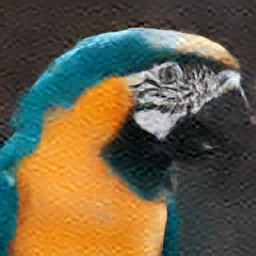} &
\includegraphics[width=1.0\linewidth]{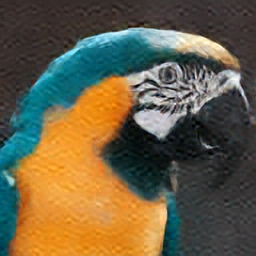} 
\end{tabular}
\vspace{-.7em}
\caption{\textbf{(Top)}: \textbf{Training loss} of $\x$- and $\v$-prediction, using the same loss space of $\v$-loss (\cref{tab:3x3s}(a), third row). We plot the loss averaged per pixel per channel.
\textbf{(Bottom)}: \textbf{Denoised images} from $\x$- and $\v$-prediction, where $\v$-prediction's denoised output is visualized according to \cref{tab:xev}(c)(1). The denoised image from $\v$-prediction has noticeable artifacts, as reflected by the higher loss.
}
\label{fig:loss_curves}
\end{figure}

\subsection{Training Loss and Denoised Images}

In \cref{tab:3x3s}(a), the failure of $\e$-/$\v$-prediction is caused by the inherent incapability of predicting a high-dimensional output from a limited-capacity network. This failure can be seen from the training loss curves.

In \cref{fig:loss_curves}\,(top), we compare the training loss curves under the same $\v$-loss, defined as $\mathcal{L}=\mathbb{E} \| \v_\theta(\z_t, t) - \v \|^2$, using $\v$-prediction (\ie, $\v_\theta=\net_\theta$) versus $\x$-prediction (\ie, $\v_\theta=(\net_\theta - \z_t)/(1-t)$). Since the loss is computed in the same space and only the parameterization differs, comparing the loss values is legitimate.

\cref{fig:loss_curves}\,(top) shows that $\v$-prediction yields substantially higher loss values (about 25\%) than $\x$-prediction, even though $\v$-prediction appears to be the ``native'' parameterization for the $\v$-loss. This comparison indicates that the task of $\x$-prediction is inherently \textit{easier}, as the data lie on a low-dimensional manifold. We also observe that $\e$-prediction (not shown here) has about $3\times$ higher loss and is unstable, which may explain its failure in \cref{tab:3x3s}(a).

\cref{fig:loss_curves}\,(bottom) compares the \textit{denoised} images corresponding to the two training curves. For $\x$-prediction, the denoised image is simply $\x_\theta=\net_\theta$; for $\v$-prediction, the denoised image is $\x_\theta{=}(1{-}t){\net_\theta}{+} \bm{z}_t$ with $\v_\theta{=}\net_\theta$ (see \cref{tab:xev}(c)(1)). The \textit{higher} loss of $\v$-prediction in \cref{fig:loss_curves}\,(top) can be reflected by its noticeable \textit{artifacts} in \cref{fig:loss_curves}\,(bottom).

Note that the artifact in \cref{fig:loss_curves}\,(bottom) is that of a \textit{single} denoising step. In the generation process, this error can accumulate in the multi-step ODE solver, which leads to the catastrophic failure in \cref{tab:3x3s}(a).

\subsection{Pre-conditioner}

In EDM \cite{Karras2022edm}, an extra ``pre-conditioner'' is applied to wrap the direct output of the network. Using the notation of our paper, the pre-conditioner formulation can be written as:
$\x_\theta(\z_t, t) = c_\text{skip} \cdot \z_t + c_\text{out}\cdot \net_\theta(\z_t, t)$,
where $c_\text{skip}$ and $c_\text{out}$ are pre-defined coefficients.
This equation suggests that \textit{unless} $c_\text{skip}\equiv0$, the network output in a pre-conditioner must \textit{not} perform $\x$-prediction. And according to the manifold assumption, this formulation should not remedy the issue we consider, as we examine next.

\paragraph{Formulation of pre-conditioners.}
Given the definition of pre-conditioner, we can rewrite the ``pre-conditioned $\x$-prediction'' as:
\begin{equation}
\left\{
\begin{aligned}
\x_\theta &= c_\text{skip} \cdot \z_t + c_\text{out}\cdot \net_\theta \\
\e_\theta &= (\z_t - t  \x_\theta) / (1 - t)  \\
\v_\theta &= (\x_\theta - \z_t) / (1 - t)
\end{aligned}
\right.
\end{equation}
Accordingly, the objective in \cref{eq:objective} ($\v$-loss) is written as:
\begin{gather}
\mathcal{L} = \mathbb{E}_{t,\x,\e} \Big\|  \v_\theta(\z_t, t) - \v \Big\|^2,
 \\
\text{where:}\quad \v_\theta(\z_t, t) = (\x_\theta(\z_t, t)-\bm{z}_t) / (1-t) \nonumber \\
\text{and}
\quad \x_\theta(\z_t, t) = c_\text{skip} \cdot \z_t + c_\text{out}\cdot \net_\theta(\z_t, t). \nonumber
\end{gather}
Comparing with \cref{eq:objective}, the only difference is in how we compute $\x_\theta$ from the network.

\begin{table}[t]
\centering
\tablestyle{2.5pt}{1.2}
\begin{tabular}{x{36} | x{56} | x{56} |  x{56} } 
&  & \multicolumn{2}{c}{\textbf{pre-conditioned} predictions} \\
& \textbf{$\x$-pred}  & EDM-style & linear \\
\cline{2-4}
 & $c_\text{skip}=0$
 & $c_\text{skip}{=}\frac{1}{t}\frac{\sigma_\text{data}^2}{\sigma_\text{data}^2 + \sigma_t^2}$
 & $c_\text{skip}=t$\\
 & $c_\text{out}=1$
 & $c_\text{out}{=}\frac{\sigma_\text{data}\sigma_t}{\sqrt{\sigma_\text{data}^2 + \sigma_t^2}}$
 & $c_\text{out}=1-t$\\
\shline
$\x$\textbf{-loss} & \cellcolor{good} {10.14} & \cellcolor{bad} 28.94  & \cellcolor{bad} 39.50 \\
\hline
$\eps$\textbf{-loss} & \cellcolor{good} {10.45} & \cellcolor{bad} 72.05  & \cellcolor{bad}  67.56 \\
\hline
$\v$\textbf{-loss} & \cellcolor{good} \hspace{2pt} {8.62} & \cellcolor{bad} 35.49 & \cellcolor{bad} 46.25 \end{tabular}
\vspace{.0em}
\caption{\textbf{Comparisons with pre-conditioners} (FID-50K, ImageNet 256, JiT-B/16). The settings are the same as \cref{tab:3x3s} (a).
}
\label{tab:precond}
\end{table}

As EDM \cite{Karras2022edm} uses a variance-exploding schedule (that is, $\z_t{=}\x{+}\sigma_t\e$)  and we use a (roughly) variance-preserving version (that is, $\z_t{=}t\x{+}(1{-}t)\e$), a fully equivalent conversion is impossible. 
To have a pre-conditioner in our case, we rewrite our version as $
\frac{1}{t} \z_t = \x + \frac{1 - t}{t}\e
$. As such, we set $\sigma_t = \frac{1-t}{t}$, which is the noise added to unscaled images, similar to EDM's.
With this, we can rewrite the coefficients defined by EDM \cite{Karras2022edm} as:
$c_\text{skip}=\frac{1}{t} \frac{\sigma_\text{data}^2}{\sigma_\text{data}^2 + \sigma_t^2}$
and $c_\text{out}=\frac{\sigma_\text{data}\sigma_t}{\sqrt{\sigma_\text{data}^2 + \sigma_t^2}}$
where $\sigma_\text{data}$ is the data standard deviation set as 0.5 \cite{Karras2022edm}. We choose $c_\text{in}= t$ ($=\frac{1}{\sigma_t+1}$), so that the direct input to the network (\ie, $c_\text{in}\cdot\frac{1}{t}z_t$) is still $z_t$.
It can be shown that \textit{only} when
$t \rightarrow 0$, we have: $\sigma_t{\rightarrow}+ \infty$, $c_\text{skip}{\rightarrow}0$, $c_\text{out}{\rightarrow}1$,
which approaches $\x$-prediction.
We also consider a simpler \textit{linear} pre-conditioner:
$c_\text{skip}=t$ and $c_\text{out}=1 - t$,
which also performs $\x$-prediction \textit{only} when $t = 0$.

\paragraph{Results of pre-conditioners.}
\cref{tab:precond} shows that the pre-conditioned versions all fail catastrophically, suggesting that deviating from $\x$-prediction is not desired in high-dimensional spaces. Interestingly, the pre-conditioned versions are much better than $\e$-/$\v$-prediction in \cref{tab:3x3s}(a). We hypothesize that this is because they are more similar to $\x$-prediction when $t{\rightarrow}0$, which alleviates this issue.

\subsection{Exploration: Classification Loss}

Our paper focuses on a minimalist design, and we intentionally do \textit{not} use any extra loss. However, we note that latent-based methods \cite{Rombach2022} typically rely on tokenizers trained with \textit{adversarial} and \textit{perceptual} losses, and thus their generation process is not purely driven by diffusion. Next, we discuss a simple extension on our pixel-based models with an additional classification loss.

\begin{table}[t]
\centering
\tablestyle{6pt}{1.1}
\begin{tabular}{c | x{36} | x{36} }
 FID-50K & \textbf{JiT-B/16} & \textbf{JiT-L/16} \\
\shline
$\v$-loss only & 4.37 & 2.79 \\
w/ cls loss & \textbf{4.14} & \textbf{2.50}
\end{tabular}
\vspace{-.5em}
\caption{\textbf{Exploration: additional classification loss}. We do \textbf{\textit{not}} use this loss in any other experiments.
Settings: ImageNet 256$\times$256, 200-ep, with CFG interval.
}
\vspace{-1em}
\label{tab:cls}
\end{table}

Formally, we attach a classifier head after a specific Transformer block (the 4th in JiT-B and the 8th in JiT-L). The classifier consists of global average pooling followed by a linear layer, and a cross-entropy loss is applied for the 1000-class ImageNet classification task.
This classification loss $\mathcal{L}_\text{cls}$ is scaled by a weight $\lambda$ (\eg, $100$) and added to the $\ell_2$-based (\ie, element-wise sum of squared errors) regression loss. To prevent label leakage, we disable class conditioning for all layers before the classifier head. We note that this modification remains minimal and does not rely on any pre-trained classifier, unlike the perceptual loss \cite{Zhang2018}.
This minor modification leads to a decent improvement, as shown in \cref{tab:cls}. This exploration suggests further potential for combining our simple method with additional loss terms, which we leave for future work.

Despite the improvement, we do \textit{\textbf{not}} use this or any additional loss in the other experiments presented in this paper.

\subsection{Cross-resolution Generation}

A model trained at one resolution can be applied to another by simply downsampling or upsampling the generated images. We refer to this as \textit{cross-resolution} generation. In our setup, JiT/16 at 256 and JiT/32 at 512 have comparable parameters and compute, making their cross-resolution comparison meaningful. The results are in \cref{tab:cross}
\begin{table}[h]
\vspace{-.5em}
\centering
\tablestyle{2pt}{1.1}
\begin{tabular}{l | c }
  & FID@256 \\
\shline
\textbf{JiT-G/16}@256 &  1.82 \\
\textbf{JiT-G/32}@512, $\downarrow$256 & \cellcolor[gray]{.9}{1.84} \\
\end{tabular}
\hspace{1em}
\tablestyle{2pt}{1.1}
\begin{tabular}{l | c }
  & FID@512 \\
\shline
\textbf{JiT-G/16}@256, $\uparrow$512 & \cellcolor[gray]{.9}{{2.45}} \\
\textbf{JiT-G/32}@512 &  1.78 \\
\end{tabular}

\vspace{-.5em}
\caption{\textbf{Cross-resolution Generation} (noted in gray), using \mbox{JiT-G/16} trained at resolution 256 and JiT-G/32 trained at 512, followed by upsampling ($\uparrow$) or downsampling ($\downarrow$). All entries have similar parameters and flops (see \cref{tab:in256-sys} and \ref{tab:in512-sys}).
}
\label{tab:cross}
\vspace{-.5em}
\end{table}

\textit{Downsampling} the images generated by the 512 model to 256 resolution yields a decent FID@256 of 1.84. 
This result remains competitive when compared with the 256-resolution expert (FID@256 of 1.82), while maintaining a similar computational cost and gaining the additional ability to generate at 512 resolution.

On the other hand, \textit{upsampling} the images generated by the 256 model to 512 resolution results in a noticeably worse FID@512 of 2.45, compared with the 512-resolution expert's FID@512 of 1.78.
This degradation is caused by the loss of higher-frequency details due to upsampling.

\subsection{Additional Metrics}

For completeness, we report precision and recall \cite{Kynkaanniemi2019} on ImageNet 256$\times$256 in \cref{tab:prec-recall}, compared with the commonly used baselines of DiT and SiT, and the latest RAE:

\begin{table}[h]
\vspace{-.5em}
\tablestyle{6pt}{1.05}
\begin{tabular}{l | c | c | c c }
 & FID$\downarrow$ & IS$\uparrow$ & Prec$\uparrow$ & Rec$\uparrow$ \\
\shline
DiT-XL/2 \cite{Peebles2023} & 2.27 & 278.2 & 0.83 & 0.57 \\
SiT-XL/2 \cite{ma2024sit} & 2.06 & 277.5 & 0.82 & 0.59 \\
RAE \cite{zheng2025diffusion}, DiT$^{\text{DH}}$-XL/2 & 1.13 & 262.6 & 0.78 & 0.67 \\
\hline
JiT-B/16 & 3.66 & 275.1 & 0.82 & 0.50 \\
JiT-L/16 & 2.36 & 298.5 & 0.80 & 0.59 \\
JiT-H/16 & 1.86 & 303.4 & 0.78 & 0.62 \\
JiT-G/16 & 1.82 & 292.6 & 0.79 & 0.62 \\
\end{tabular}
\vspace{-.5em}
\caption{Precision and recall on ImageNet 256$\times$256.}
\label{tab:prec-recall}
\vspace{-.5em}
\end{table}

\section{Qualitative Results}

In \cref{fig:results1} to \ref{fig:results4}, we provide additional \textit{uncurated} examples on ImageNet 256$\times$256.

\newpage

\vspace{1em}
\paragraph{Acknowledgements.} 
We thank Google TPU Research Cloud (TRC) for granting us access to TPUs, and the MIT ORCD Seed Fund Grants for supporting GPU resources.

{\small
\bibliographystyle{configs/ieeenat_fullname}
\bibliography{jit.bib}
}

\clearpage

\renewcommand{\hhs}{\hspace{-0.001em}}
\renewcommand{\vvs}{\vspace{-.1em}}

\newcommand{\tilewidth}{0.14\linewidth}

\newcommand{\imgcaptiontext}{
\textit{Uncurated} samples on ImageNet 256$\times$256 using JiT-G/16 conditioned on the specified classes. Unlike the common practice of visualizing with a higher CFG, here we show images using the CFG value (2.2) that achieves the reported FID of 1.82.}

\newcommand{\sampledir}{samples256_cfg2.2_jpg}

\newcommand{\addclass}[2]{
\begin{minipage}[t]{0.49\linewidth}
\centering
\includegraphics[width=\tilewidth]{\sampledir/cls#1/000#1.jpg}\hhs
\includegraphics[width=\tilewidth]{\sampledir/cls#1/001#1.jpg}\hhs
\includegraphics[width=\tilewidth]{\sampledir/cls#1/002#1.jpg}\hhs
\includegraphics[width=\tilewidth]{\sampledir/cls#1/003#1.jpg}\hhs
\includegraphics[width=\tilewidth]{\sampledir/cls#1/004#1.jpg}\hhs
\includegraphics[width=\tilewidth]{\sampledir/cls#1/005#1.jpg}\hhs
\includegraphics[width=\tilewidth]{\sampledir/cls#1/006#1.jpg}\vvs
\\
\includegraphics[width=\tilewidth]{\sampledir/cls#1/007#1.jpg}\hhs
\includegraphics[width=\tilewidth]{\sampledir/cls#1/008#1.jpg}\hhs
\includegraphics[width=\tilewidth]{\sampledir/cls#1/009#1.jpg}\hhs
\includegraphics[width=\tilewidth]{\sampledir/cls#1/010#1.jpg}\hhs
\includegraphics[width=\tilewidth]{\sampledir/cls#1/011#1.jpg}\hhs
\includegraphics[width=\tilewidth]{\sampledir/cls#1/012#1.jpg}\hhs
\includegraphics[width=\tilewidth]{\sampledir/cls#1/013#1.jpg}\vvs
\\
\includegraphics[width=\tilewidth]{\sampledir/cls#1/014#1.jpg}\hhs
\includegraphics[width=\tilewidth]{\sampledir/cls#1/015#1.jpg}\hhs
\includegraphics[width=\tilewidth]{\sampledir/cls#1/016#1.jpg}\hhs
\includegraphics[width=\tilewidth]{\sampledir/cls#1/017#1.jpg}\hhs
\includegraphics[width=\tilewidth]{\sampledir/cls#1/018#1.jpg}\hhs
\includegraphics[width=\tilewidth]{\sampledir/cls#1/019#1.jpg}\hhs
\includegraphics[width=\tilewidth]{\sampledir/cls#1/020#1.jpg}\vvs
\\
{\scriptsize {class #1}: #2}
\vspace{.5em}
\end{minipage}
}

\begin{figure*}[t]
\centering
\addclass{012}{house finch, linnet, Carpodacus mexicanus}
\addclass{014}{indigo bunting, indigo finch, indigo bird, Passerina cyanea
}
\\
\addclass{042}{agama}
\addclass{081}{ptarmigan}
\\
\addclass{107}{jellyfish}
\addclass{108}{sea anemone, anemone}
\\
\addclass{110}{flatworm, platyhelminth}
\addclass{117}{chambered nautilus, pearly nautilus, nautilus}
\\
\addclass{130}{flamingo}
\addclass{279}{Arctic fox, white fox, Alopex lagopus}
\\
\caption{\imgcaptiontext}
\label{fig:results1}
\vspace{-1em}
\end{figure*}

\begin{figure*}[t]
\centering
\addclass{288}{leopard, Panthera pardus}
\addclass{309}{bee}
\\
\addclass{349}{bighorn, bighorn sheep, cimarron, Rocky Mountain bighorn}
\addclass{397}{puffer, pufferfish, blowfish, globefish}
\\
\addclass{425}{barn}
\addclass{448}{birdhouse}
\\
\addclass{453}{bookcase}
\addclass{458}{brass, memorial tablet, plaque}
\\
\addclass{495}{china cabinet, china closet}
\addclass{500}{cliff dwelling}
\\
\caption{\imgcaptiontext}
\label{fig:results2}
\vspace{-1em}
\end{figure*}

\begin{figure*}[t]
\centering
\addclass{658}{mitten}
\addclass{661}{Model T}
\\
\addclass{718}{pier}
\addclass{724}{pirate, pirate ship}
\\
\addclass{725}{pitcher, ewer}
\addclass{757}{recreational vehicle, RV, R.V.}
\\
\addclass{779}{school bus}
\addclass{780}{schooner}
\\
\addclass{829}{streetcar, tram, tramcar, trolley, trolley car}
\addclass{853}{thatch, thatched roof}
\\
\caption{\imgcaptiontext}
\label{fig:results3}
\vspace{-1em}
\end{figure*}

\begin{figure*}[t]
\centering
\addclass{873}{triumphal arch}
\addclass{900}{water tower}
\\
\addclass{911}{wool, woolen, woollen}
\addclass{913}{wreck}
\\
\addclass{927}{trifle}
\addclass{930}{French loaf}
\\
\addclass{946}{cardoon}
\addclass{947}{mushroom}
\\
\addclass{975}{lakeside, lakeshore}
\addclass{989}{hip, rose hip, rosehip}
\\
\caption{\imgcaptiontext}
\label{fig:results4}
\vspace{-1em}
\end{figure*}
\end{document}